\documentclass[10pt,twocolumn,letterpaper]{article}

\usepackage[pagenumbers]{cvpr} 

\newcommand{\spin}[1]{\rotatebox{90}{#1}}
\newcommand{\un}[1]{\underline{#1}}

\usepackage[accsupp]{axessibility}
\usepackage{graphicx}
\usepackage{subcaption}
\usepackage{adjustbox}
\usepackage{amsmath}
\usepackage{booktabs}
\usepackage{tgcursor}
\usepackage{microtype}

\usepackage{algorithm} 
\usepackage{algpseudocode} 
\usepackage{listings} 
\usepackage{xcolor} 
\usepackage{multirow}   
\usepackage{rotating}
\usepackage[table]{xcolor}
\lstdefinestyle{Pytorch}{
    language         = Python,
    backgroundcolor  = \color{backgray},
    basicstyle = \fontsize{8.0pt}{9pt}\selectfont\ttfamily\bfseries,
    columns          = fullflexible,
    breaklines       = true,
    captionpos       = b,
    commentstyle     = \fontsize{4pt}{4pt}\color{JungleGreen},
    keywordstyle     = \fontsize{4pt}{4pt}\color{codekw},
    morekeywords     = {augment, softmax, PGD, L_clip, torch, argmax, argmin, t, log, norm, mean, backward, step},
}
\definecolor{codeblue}{rgb}{0.15, 0.9, 0.}
\definecolor{codekw}{rgb}{0.35, 0.35, 0.9}
\definecolor{backgray}{gray}{0.95}

\definecolor{cvprblue}{rgb}{0.21,0.49,0.74}
\usepackage[pagebackref,breaklinks,colorlinks,allcolors=cvprblue]{hyperref}

\def\paperID{****} 
\def\confName{CVPR}
\def\confYear{2026}

\title{Finetune Like You Pretrain: Boosting Zero-shot Adversarial Robustness in Vision-language Models}

\author{
Songlong Xing$^{1}$\thanks{
Work done during visit to Torr Vision Group, University of Oxford.
} \quad
Weijie Wang$^{1,2}$\thanks{
Corresponding author
}
\quad
Zhengyu Zhao$^{3}$ \quad
Jindong Gu$^{4}$ \quad
Philip Torr$^{4}$ \quad
Nicu Sebe$^{1}$ \\[0.4em]
$^{1}$ University of Trento, Italy \quad
$^{2}$ Fondazione Bruno Kessler, Italy \\
$^{3}$ Xi'an Jiaotong University, China \quad
$^{4}$ University of Oxford, UK \\
{\tt\small
\{songlong.xing, weijie.wang\}@unitn.it
}
}

\begin{document}
\maketitle
\begin{abstract}
Despite their impressive zero-shot abilities, vision-language models such as CLIP have been shown to be susceptible to adversarial attacks. To enhance its adversarial robustness, recent studies finetune the pretrained vision encoder of CLIP with adversarial examples on a proxy dataset such as ImageNet by aligning adversarial images with correct class labels. However, these methods overlook the important roles of training data distributions and learning objectives, resulting in reduced zero-shot capabilities and limited transferability of robustness across domains and datasets. In this work, we propose a simple yet effective paradigm AdvFLYP, which follows the training recipe of CLIP's pretraining process when performing adversarial finetuning to the model. Specifically, AdvFLYP finetunes CLIP with adversarial images created based on image-text pairs collected from the web, and match them with their corresponding texts via a contrastive loss. To alleviate distortion of adversarial image embeddings of noisy web images, we further propose to regularise AdvFLYP by penalising deviation of adversarial image features. We show that logit- and feature-level regularisation terms benefit robustness and clean accuracy, respectively. Extensive experiments on 14 downstream datasets spanning various domains show the superiority of our paradigm over mainstream practices. Our code is available \href{https://github.com/Sxing2/AdvFLYP}{here}.

\end{abstract}    
\section{Introduction}
\label{sec:intro}

\begin{figure}[t]
    \centering
    \begin{subfigure}{0.8\linewidth}
        \centering
        \includegraphics[width=\linewidth]{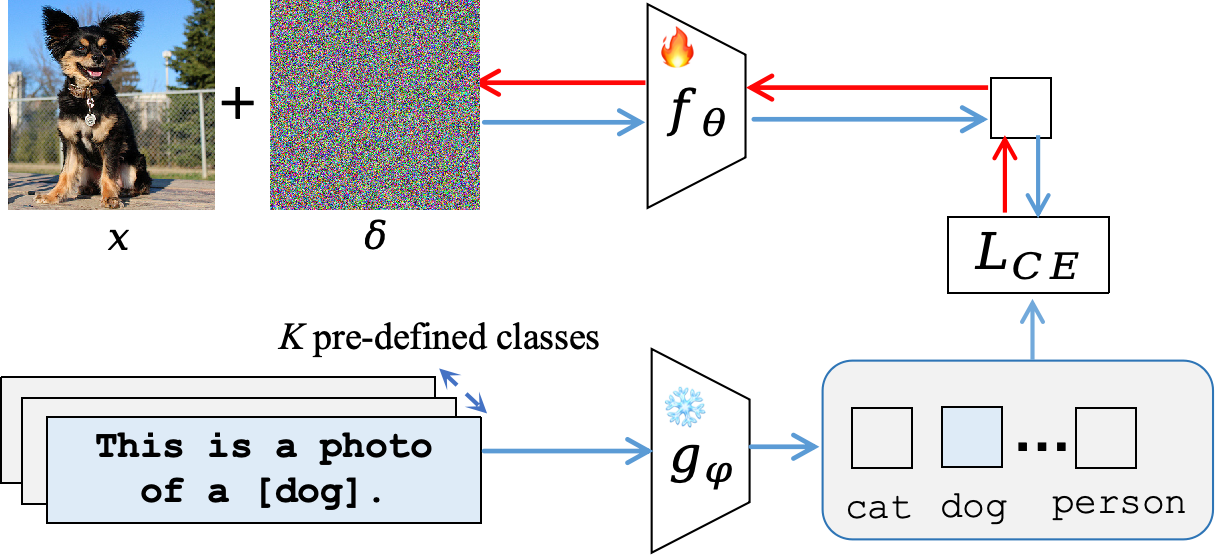}
        \caption{Mainstream adversarial finetuning paradigm.}
        \label{fig:curr_AFT}
    \end{subfigure}
    
    \adjustbox{vspace=.1em}{}

    \begin{subfigure}{0.9\linewidth}
        \centering
        \includegraphics[width=\linewidth]{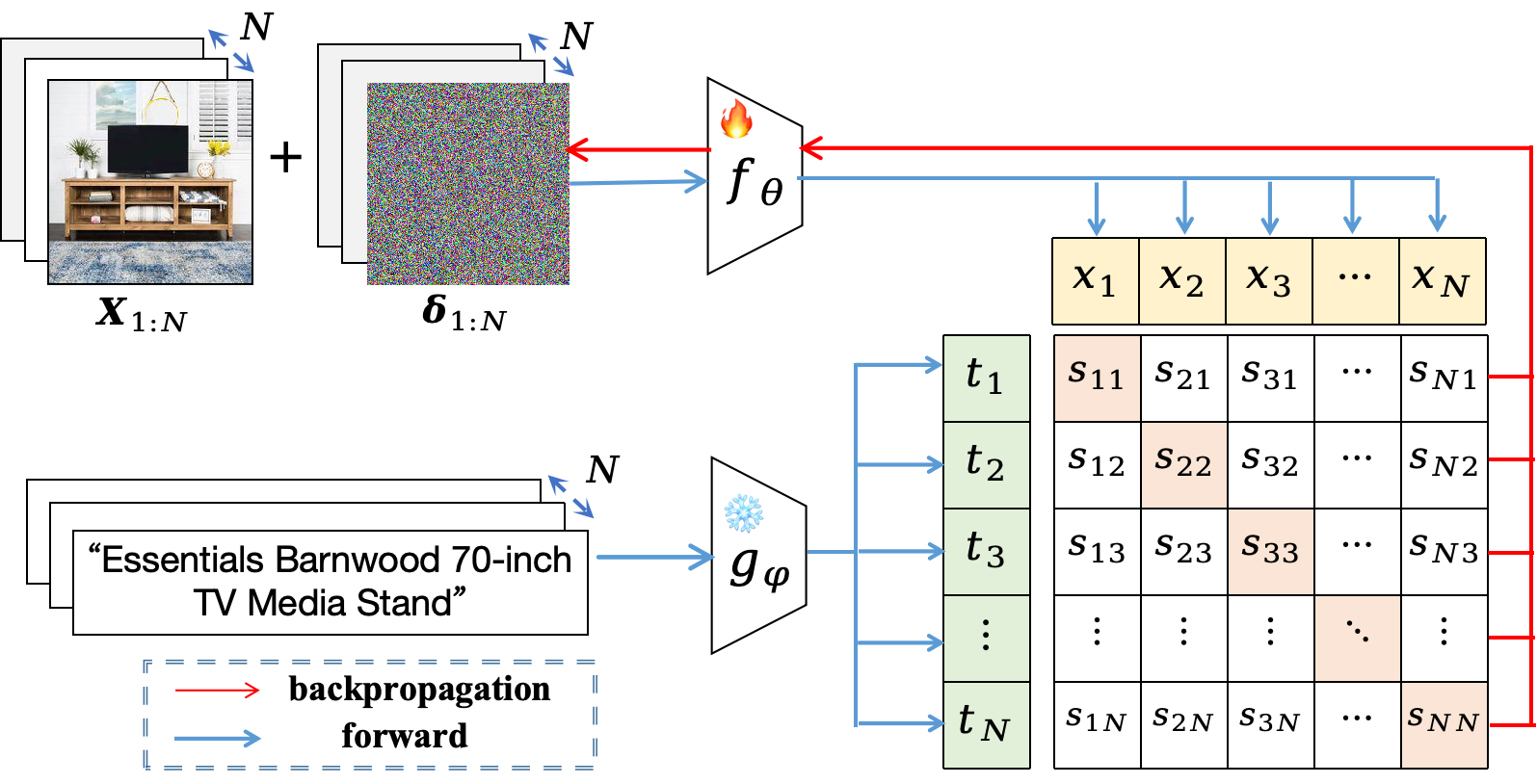}
        \caption{An overview of our AdvFLYP paradigm.}
        \label{fig:AdvFLYP}
    \end{subfigure}
    \caption{A basic illustration of current mainstream AFT methods and our AdvFLYP paradigm.
    Rather than performing AFT on a well-curated proxy dataset via a cross-entropy loss, AdvFLYP adversarially finetunes the vision encoder on web-scale image-text pairs via a contrastive loss.
    }\label{fig:teaser}
    \vspace{-2em}
\end{figure}

Vision-language models (VLMs) bridge the gap between visual and language modalities by pretraining the models over web-scale image-text data \cite{radford2021learning,jia2021scaling,li2022blip,li2023blip,liu2023visual,zhuminigpt}. 
As a notable representative, CLIP \cite{radford2021learning} leverages a dual-encoder architecture to encode vision and language into the same latent space, where cosine similarity between the embeddings of an image and a text can be computed. 
Having been trained to match images with their descriptive texts via a contrastive loss, CLIP possesses remarkable amounts of real-world knowledge, which can be leveraged for zero-shot inference in downstream tasks \citep{pratt2023does,saha2024improved,sammani2024interpreting}.
Despite its widespread deployment in numerous scenarios \cite{zhong2022regionclip,wang2025declip,zhang2022pointclip,luddecke2022image,patashnik2021styleclip},
recent studies have revealed its alarming vulnerability to adversarial attacks \cite{mao2023understanding};
A slight pixel-level perturbation added onto the image at inference time can mislead the model into making wrong predictions \cite{szegedy2014intriguing,zhang2019theoretically}.

Prior studies on non-foundational deep models establish adversarial training (AT) as the most effective method to train a robust model from scratch \cite{madry2018towards,Pan_Li_Yao_2024}, which alternately creates adversarial images and employs them to train the model.
Built upon this idea, recent work on robustifying CLIP leverages a proxy classification dataset, such as ImageNet \cite{deng2009imagenet} and TinyImageNet \cite{le2015tiny}, and finetunes the vision encoder of CLIP to align adversarial images with their correct class labels (\cref{fig:curr_AFT}) \cite{mao2023understanding,wang2024pre,yu2024text,waseda2025quality}.
This adversarial finetuning (AFT) process leads to CLIP models that are robust to adversarial attacks on a spectrum of downstream tasks without the need for further training, achieving \textit{zero-shot adversarial robustness} \cite{mao2023understanding}.
However, these methods cause a noticeable degradation of the zero-shot capabilities in CLIP \cite{xing2025clip}, and exhibit limited transferability of robustness on cross-domain data.
Intuitively, AFT for strengthening adversarial robustness of CLIP differs from conventional AT in the following ways.
Firstly, CLIP has learned large amounts of real-world knowledge through language supervision, and modifying its model weights, even based on clean images, degrades generalisation of the model.
Secondly, different from conventional models, CLIP has been pretrained to match images with corresponding texts, rather than image classification. 
Therefore, finetuning CLIP on a classification dataset, albeit reasonable in the sense that the finetuned model is to be evaluated on downstream classification datasets at test time, drastically deviates from the pretraining process.
Additionally, leveraging a cross-entropy loss in AFT risks compromising the capability of the visual encoder because multiple images from the same class are aligned with a single text prompt (\eg, `\textit{a photo of a dog}').

Recent research on \textit{robust finetuning}
\footnote{
The aim of \textit{robust finetuning} is to improve generalisation of the model on OOD data, and should not be confused with adversarial robustness.
}
finds that sticking to the same contrastive objective as employed in CLIP's pretraining process in the course of finetuning leads to better performance on both \textit{in-distribution (ID)} and \textit{out-of-distribution (OOD)} data, compared to the straightforward cross-entropy loss \cite{goyal2023finetune}.
In this work, we draw inspiration from this finding and consider the training data and the training objective in AFT to mitigate existing limitations as discussed above.
To this end, we propose a simple yet effective AFT paradigm for CLIP, termed \textit{Adversarially Finetune Like You Pretrain} (AdvFLYP), which inherits the training recipe of the pretraining process of CLIP when performing adversarial finetuning.
Specifically, we alternately create adversarial images based on image-text data collected from the web, and align these adversarial images with their corresponding texts (\cref{fig:AdvFLYP}).
Previous research on AFT finds that logit-level regularisation of the adversarial images guided by the original CLIP facilitates robustness and generalisation \cite{wang2024pre}.
In this work, although web-scale image-text pairs simulate CLIP's pretraining data with diverse coverage, they are generally noisy and can cause distorted features of adversarial images, which risks hurting the vision encoder.
Therefore, we propose a feature-level regulariser, which penalises the deviation of the adversarial image features from the clean counterparts in the embedding space.
We find that logit- and feature-level regularisation terms benefit downstream robustness and preservation of zero-shot capabilities, repsectively.
Experiments show that AdvFLYP outperforms prior AFT methods on 14 downstream classification datasets spanning a wide spectrum of domains, even though our AFT process does not cater to the classification task, but rather respects the contrastive training recipe of CLIP's pretraining.

Our contributions are summarised as follows:
\begin{itemize}
    \item This work proposes a simple yet effective AdvFLYP paradigm for adversarial finetuning of CLIP, which respects the pretraining process of CLIP, challenging existing practices where CLIP is finetuned to classify adversarial images correctly.
    \item We propose to regularise AdvFLYP by penalising the the deviation of the adversarial image features in the latent space. We show that logit- and feature-level regularisation benefits robustness and clean accuracy, respectively.
    \item Experiments on 14 downstream datasets show that AdvFLYP outperforms current AFT methods, establishing a new generic AFT paradigm. We hope this work raises to the community the importance of respecting the pretraining process in adversarial finetuning of VLMs.
\end{itemize}
\section{Related Work}
\label{sec:related_work}

\subsection{Adversarial Robustness of VLMs}
The susceptibility of deep networks to imperceptible adversarial noises has been widely studied \cite{carlini2017towards,szegedy2014intriguing}, since their early emergence \cite{krizhevsky2012imagenet,he2016deep}.
To defend neural models from such attacks, adversarial training (AT) has been established as the \textit{de facto} standard to train adversarially robust models from scratch \cite{madry2018towards,zhang2019theoretically,rice2020overfitting}.
In recent years, vision-language models (VLMs) such as CLIP \cite{radford2021learning} have shown excellent zero-shot abilities.
However, recent studies reveal their concerning vulnerabilities to adversarial attacks \cite{mao2023understanding,zhao2023evaluating}.
This work focuses on CLIP, while retains the possibility of applying our generic paradigm to other VLMs.
To enhance the adversarial robustness of CLIP, existing efforts fall into three main categories.
\textbf{Adversarial finetuning (AFT)} methods aim to finetune the pretrained vision encoder of CLIP on adversarial images generated on the fly.
\citet{mao2023understanding} propose TeCoA, which finetunes the model to classify adversarial images into their correct category labels via a cross-entropy loss. 
Subsequent AFT methods are mostly based on this work, with the aim to further improve robustness and mitigate clean accuracy degradation, by introducing additional regularisation terms \cite{wang2024pre,yu2024text}, improving the quality of adversarial examples \cite{dong2025improving}, and improving the text quality \cite{waseda2025quality}.
Specifically, \citet{wang2024pre} propose to regularise the model by imposing a KL-divergence loss guided by the original CLIP.
\citet{yu2024text} introduce text-guided attention to encourage the model to attend to the correct areas of adversarial images.
\citet{dong2025improving} improve the quality of adversarial examples by considering the adversarial trajectory when generating 
adversarial examples.
\citet{waseda2025quality} improve the text quality by generating semantically rich descriptions for training images using a generative VLM.
These models invariably involve a proxy classification dataset to provide original images and their correct labels.
However, this paradigm largely overlooks the important roles of training data distributions and training objectives, 
significantly deviating from CLIP's pretraining behaviour.
In this work, we highlight the importance of the training recipe and propose a simple yet effective paradigm where CLIP's pretraining behaviour is respected, achieving improved transferability of robustness and better clean accuracy retention.
This work is orthogonal to the advancements subsequent to TeCoA \cite{mao2023understanding}. 
\textbf{Adversarial prompt tuning (APT)} aims to adapt CLIP with learnable prompts to align adversarial images with ground-truth labels \cite{li2024one,zhang2024adversarial,zhou2024few}. This type of methods is based on prompt tuning of CLIP \cite{zhou2022learning,zhou2022conditional}.
More recently, \textbf{test-time defence} methods have also garnered research attention, which seek to strengthen adversarial robustness of CLIP at test time without training the model \cite{wang2025tapt,sheng2025r,xing2025clip,tong2025zero,zhang2025clipure}.
This work focuses on AFT because it is still the most straightforward and effective method to robustify VLMs.

\subsection{Robust Finetuning of VLMs}
\textit{Robust finetuning} aims to improve the robustness of the finetuned model to distribution shifts \cite{mao2024context}.
Previous work on \textit{robust finetuning} shows that subtle changes to the finetuning procedure have significant impacts on the performance on out-of-distribution (OOD) data \cite{kumar2022finetuning}.
To improve robustness to distribution shifts while maintaining high performance on in-distribution (ID) data, numerous finetuning methods have been proposed \cite{mao2024context,goyal2023finetune,oh2024towards,nam2024lipsumft}.
In \textit{adversarial finetuning} on which this work focuses, the aim is to achieve zero-shot adversarial robustness of the model on various downstream datasets.
However, the `robust finetuning' in an adversarial context is still understudied, with existing AFT efforts improving robustness based on a classification dataset without rethinking the training recipe.
The most inspiring to our work is \textit{Finetune Like You Pretrain (FLYP)} \cite{goyal2023finetune}, which shows that 
employing the contrastive loss as utilised for pretraining during finetuning outperforms methods that directly leverage a standard cross-entropy loss for finetuning.
In this work, we show that respecting the training recipe of CLIP's pretraining in \textit{adversarial finetuning} significantly improves robustness across downstream datasets and domain shifts, establishing a simple yet effective AFT paradigm.

\section{Method}
\label{sec:method}

This section introduces preliminaries regarding CLIP and AFT, and elaborates on our paradigm, termed \textit{Adversarially Finetune Like You Pretrain} (AdvFLYP).

\subsection{Preliminaries}

CLIP \cite{radford2021learning} is a dual-encoder architecture with a vision encoder $f_\theta(\cdot)\in\mathbb{R}^d$ and a text encoder $g_\phi(\cdot)\in\mathbb{R}^d$, which encode images and texts into embeddings in the same latent space.
In the pretraining process, $f_\theta(\cdot)\in\mathbb{R}^d$ and $g_\phi(\cdot)\in\mathbb{R}^d$ are trained on over 400 million web-scale image-text pairs via a contrastive loss \citep{oord2018representation} to match images with their corresponding texts.
Given a batch of image-text pairs $\{(x_i,t_i)\}_{i=1}^N$, the cosine similarity between an image $x_i$ and a text $t_j$ is computed, \ie, $s_{ij}=\frac{f_{\theta}(x_i)^\intercal g_\phi(t_j)}{\parallel f_{\theta}(x_i)\parallel\parallel g_\phi(t_j)\parallel}$.
The contrastive loss is then formulated as follows:
\begin{equation}\label{eq:clip_contrastive}
\begin{split}
    &\mathcal{L}_{CLIP}
    \left(
    \{(x_i,t_i)\}_{i=1}^N \,\vert\, \theta,\phi
    \right) = \\
    & -\frac{1}{2N}\sum_{i=1}^{N}\left[\log\frac{\exp(s_{ii}/\tau)}{\sum_{j=1}^{N}\exp(s_{ij}/\tau)}+\log\frac{\exp(s_{ii}/\tau)}{\sum_{j=1}^{N}\exp(s_{ji}/\tau)}\right]
\end{split}
\end{equation}
where $\tau$ is the temperature value.
After the pretraining process, CLIP possesses the ability to perform image classification in a zero-shot manner. 
At inference time, given a downstream classification dataset with a set of $K$ pre-defined textual categories $\{c_1,\dots c_K\}$, CLIP classifies an image $x_{test}$ as the category that has the highest cosine similarity:
\begin{equation}
    \hat{y}=\arg \max_{k}\frac{f_\theta(x_{test})^\intercal g_\phi(T[c_k])}{\|f_\theta(x_{test})\|\cdot\|g_\phi(T[c_k]\|}
\end{equation}
where $T[\cdot]$ is a textual template, which is usually \textit{`This is a photo of a [CLS]'}.

\noindent
\textbf{Adversarial attacks.}
CLIP is highly vulnerable to adversarial attacks, meaning that a slight pixel-level perturbation $\delta_{adv}$
can be manipulated to maximise the classification (cross-entropy) loss. 
For example, given a test image $x_{test}\in\mathbb{R}^{C\times H\times W}$ with the ground-truth label $c_T$ from a classification dataset with $K$ classes $\{c_1,\dots,c_K\}$, the cross-entropy loss is computed as follows:
\begin{equation}\label{eq:ce_loss}
   \mathcal{L}_{ce}\left(f_\theta(x_{test})\right) = -\log \frac {\exp(s_T)} {\sum_{k=1}^K \exp(s_k)}
\end{equation}
where 
$s_k = \frac
{f_\theta(x_{test})^\intercal g_\phi(T[c_k])}
{\|f_\theta(x_{test})\| \cdot \|g_\phi(T[c_k])\|}$
is the cosine similarity of $x_{test}$ to class $c_k$.
The perturbation $\delta\in\mathbb{R}^{C\times H\times W}$ is therefore optimised to maximise this loss:
\begin{equation}\label{eq:tecoa_inner_loop}
    \delta_{adv} = \arg\max_{\delta} \mathcal{L}_{ce}(
    f_\theta\left( x_{test}+\delta \right)
    ),\;\; s.t.\, \|\delta\|_{\infty}\leq \epsilon
\end{equation}
where $\epsilon$ is the attack budget which is usually very small (\eg, $1/255$).
This process can be approximated by the PGD attack algorithm \cite{carlini2017towards}.
The adversarial image is the addition of the original image and the perturbation, \ie, $x_{adv}:=x+\delta_{adv}$.

\noindent
\textbf{Adversarial finetuning (AFT)} is a straightforward approach to robustifying CLIP by finetuning the vision encoder $f_\theta$, based on the \textit{adversarial training} (AT) \cite{madry2018towards} principles. 
As in AT, given natural training images $\textbf{X}=\left\{x^1,\dots,x^N\right\}$, AFT of CLIP alternately create adversarial images $\textbf{X}_{adv}=\left\{x^1_{adv},\dots,x^N_{adv}\right\}$ on the fly and employ them to update the model weights $\theta$. 
\citet{mao2023understanding} propose TeCoA, a fundamental paradigm that performs AFT on ImageNet,
which produces adversarial images $\textbf{X}_{adv}$ based on the cross-entropy loss (\cref{eq:tecoa_inner_loop}), and 
finetunes $f_\theta$ to correctly classify them by minimising the cross-entropy loss:
\begin{equation}\label{eq:tecoa_outer_loop}
    \theta'=\arg\min_{\theta} \frac{1}{N} \sum_{i}^N\mathcal{L}\left(f_\theta(x^i_{adv})\right)
\end{equation}
We illustrate this paradigm in \cref{fig:curr_AFT}. 
Subsequent advancements of AFT are largely based on this paradigm.
Note also that each adversarial image is produced to maximise the cross-entropy loss \wrt the class labels (\cref{eq:tecoa_inner_loop}), independently of other images in the same batch.
Although it is reasonable to involve a proxy classification dataset because the model is to be evaluated on downstream classification tasks, this paradigm deviates significantly from the pretraining process, resulting in reduced capabilities of the vision encoder and limited transferability of robustness.

\subsection{AdvFLYP}\label{sec:AdvFLYP}

This section elaborates on our AFT paradigm AdvFLYP, which highlights the importance of considering the pretraining process of VLMs when performing AFT. 

\noindent
\textbf{Data preparation.}
The training data distribution plays an important role, which is largely overlooked in previous AFT methods. 
Intuitively, utilising the pretraining data of CLIP instead of a classification dataset for AFT better retains the zero-shot capabilities of the pretrained model. 
However, since CLIP's pretraining data is not publicly available, we collect one million image-text pairs from the web to imitate a similar data distribution.
Specifically, we randomly sample one million entries with reachable URLs from LAION-400M \citep{schuhmann2021laion}.
We limit our training data amount to 1M to ensure a similar number of training images with previous methods, which employ ImageNet \cite{deng2009imagenet} with over 1.2M images for adversarial finetuning.
In our experiments, we will show that increasing the amount of image-text data from the web steadily improves the AFT performance.
In this work, we fix the dataset size at 1M to reduce training time.
We provide more information and experiments regarding the impact of training data on finetuning in Appendix~\cref{App_sec:training_data}.

\subsubsection{Training Framework}
The finetuning paradigm involves a min-max optimisation process, as in general adversarial training (AT) frameworks \cite{madry2018towards}.
We propose to leverage the contrastive objective (\cref{eq:clip_contrastive}) as employed for pretraining in our AFT paradigm.
Specifically, given a batch of image-text pairs $\{(x_i,t_i)\}_{i=1}^N$, 
we create a perturbation $\delta_i$ for each image such that the contrastive loss (\cref{eq:clip_contrastive}) within this batch is maximised:
\begin{equation}\label{eq:flyp_attack}
\begin{split}
    \boldsymbol{\delta}_{adv} = \arg&\max_{\{\delta_1,\dots,\delta_N\}}  
    \mathcal{L}_{CLIP}
    \left(\{(x_i+\delta_i,t_i)\}_{i=1}^N 
    \,\boldsymbol{\vert}\, \theta,\phi
    \right), 
    \\
    & 
    s.t.\, \left\{
    \|\delta_i\|_{\infty}\leq \epsilon \, \boldsymbol{\vert} \,i=1,\dots,N
    \right\}
\end{split}
\end{equation}
Note that the perturbations 
$\boldsymbol{\delta}_{adv}=
\left[
\delta_1^{adv},\dots,\delta_N^{adv}
\right]
\in \mathbb{R}^{ N \times C \times W \times H}
$ 
are optimised jointly, rather than independently as in previous methods.
With this batch of adversarial images, we finetune the vision encoder $f_\theta$ with the contrastive objective:
\begin{equation}\label{eq:AdvFLYP_outer_basic}
    \theta' = \arg\min_{\theta}
    \mathcal{L}_{CLIP} \left(
    \{(x_i+\delta_i^{adv},t_i)\}_{i=1}^N \,\boldsymbol{\vert}\, \theta,\phi
    \right)
\end{equation}
The finetuning process trains CLIP to match adversarial images with their corresponding texts by maximising the cosine similarity of each image with its text while treating other images as negative samples.
This is in line with CLIP's pretraining process.
\cref{fig:AdvFLYP} illustrates the basic paradigm of AdvFLYP.

\begin{figure*}[t]
    \centering
    \includegraphics[width=0.75\linewidth]{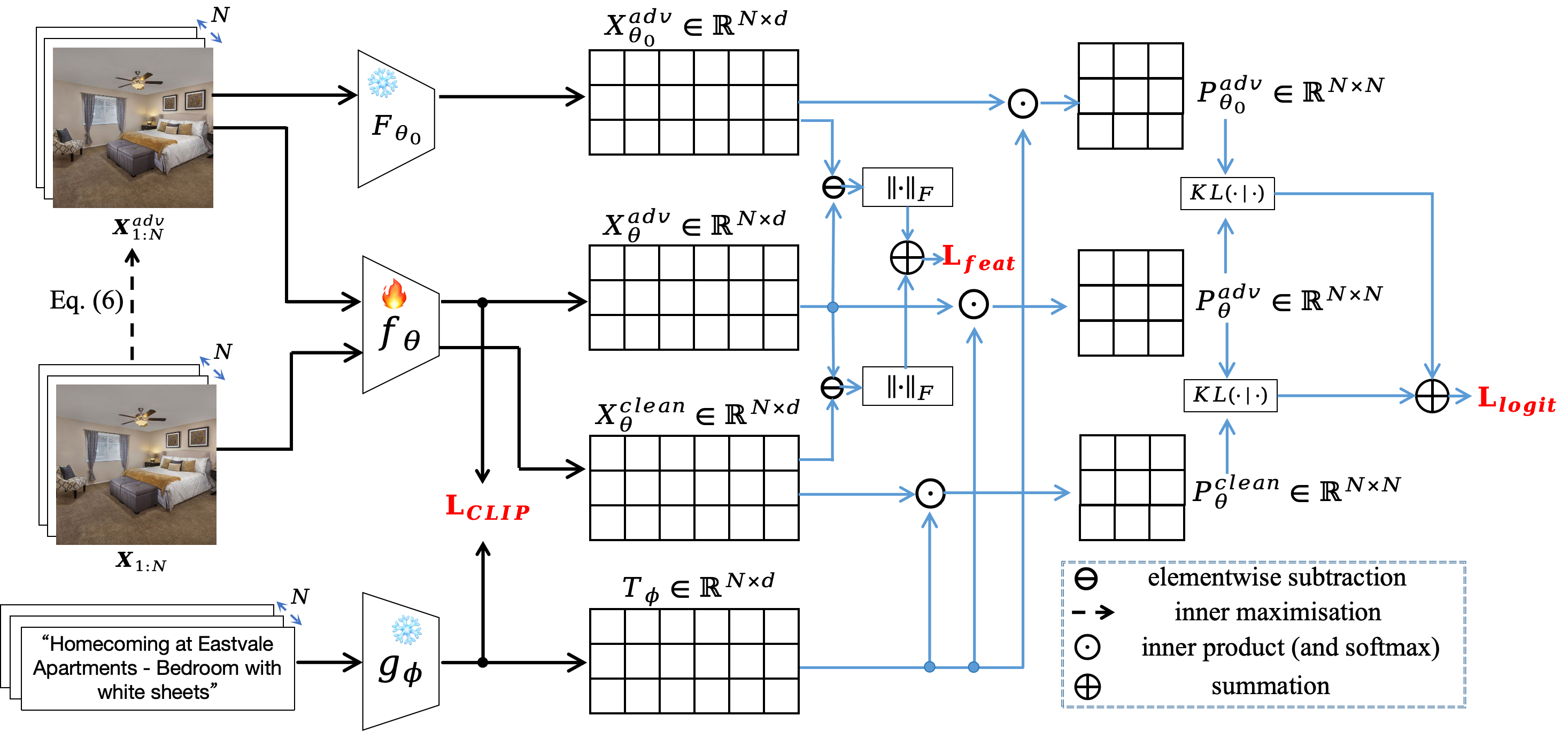}
    \caption{Overview of the formulation of $\mathcal{L}_{CLIP}$, logit-level regularisation $\mathcal{L}_{logit}$ and feature-level regularisation $\mathcal{L}_{feat}$.
    $\mathcal{L}_{CLIP}$ is the main loss of AdvFLYP (\cref{eq:AdvFLYP_outer_basic}). 
    $\mathcal{L}_{logit}$ and $\mathcal{L}_{feat}$ are only employed in the regularised variant of AdvFLYP (\cref{eq:final_loss}), 
    denoted as $\textrm{AdvFLYP}_{full}$.
    }
    \label{fig:AdvFLYP_reg}
    \vspace{-1em}
\end{figure*}

\subsubsection{Regularisation}
CLIP-guided regularisation \cite{wang2024pre,yu2024text} has been shown to improve generalisation on top of the cross-entropy loss (\cref{eq:tecoa_outer_loop} \cite{mao2023understanding}).
Specifically, \citet{wang2024pre} propose to penalise the KL-divergence \wrt to outputs by the original CLIP on the logit level.
In our AdvFLYP, we leverage web-scale image-text pairs, which are diverse but noisy, compared to labelled images from a well-curated dataset.
Creating the adversarial images by maximising the contrastive loss (\cref{eq:flyp_attack}) can lead to distorted image features in the embedding space, which can hurt the vision encoder if it is finetuned on these features.
Intuitively, imposing a penalty term that penalises the deviation of the image features output by $f_\theta$ from the clean counterparts and from the original CLIP is beneficial in preserving the capability of $f_\theta$.
Specifically, given the batch of adversarial images,
we feed them into the target model $f_\theta(\cdot)$ and the frozen original vision encoder $F_{\theta_0}(\cdot)$, \ie, 
$X_\theta^{adv}=\left[
\frac{f_\theta(x_i+\delta_i)}{\|f_\theta(x_i+\delta_i)\|}
\right]_{i=1}^N\in\mathbb{R}^{N\times d}$ 
and 
$X_{\theta_0}^{adv}=\left[
\frac{F_{\theta_0}(x_i+\delta_i)}{\|F_{\theta_0}(x_i+\delta_i)\|}
\right]_{i=1}^N\in\mathbb{R}^{N\times d}$.
We also feed the clean images to the target model and obtain their embeddings $X_{\theta}^{clean}=\left[
\frac{f_{\theta}(x_i)}{\|f_{\theta}(x_i)\|}
\right]_{i=1}^N\in\mathbb{R}^{N\times d}$.
We propose to regularise AdvFLYP on the feature level:
\begin{equation}\label{eq:feat_reg}
    \mathcal{L}_{feat}=\frac{1}{N}\left[
        \|X_\theta^{adv}-X_{\theta_0}^{adv}\|_F + \|X_\theta^{adv}-X_\theta^{clean}\|_F
    \right] 
\end{equation}
where $\|\cdot\|_F$ denotes Frobenius norm.
We also introduce the logit-level regularisation \cite{wang2024pre} in our AdvFLYP.
Specifically, given these embeddings, the probability logits of $X_\theta^{adv}$, $X_{\theta_0}^{adv}$ and $X_\theta^{clean}$ \wrt the frozen text features 
$T_{\phi}=\left[
    \frac{g_{\phi}(t_i)}{\|g_{\phi}(t_i)\|}
\right]_{i=1}^N\in\mathbb{R}^{N\times d}$ are computed as follows:
\begin{equation}
    P_\theta^{adv}=\mathrm{softmax}(X_\theta^{adv}T^\intercal)\in\mathbb{R}^{N\times N}
\end{equation}
\begin{equation}
    P_{\theta_0}^{adv}=\mathrm{softmax}(X_{\theta_0}^{adv}T^\intercal)\in\mathbb{R}^{N\times N}
\end{equation}
\begin{equation}
    P_\theta^{clean}=\mathrm{softmax}(X_\theta^{clean}T^\intercal)\in\mathbb{R}^{N\times N}
\end{equation}
The logit-level regularisation term is formulated as follows:
\begin{equation}\label{eq:logit_reg}
    \mathcal{L}_{logit}=\frac{1}{N}\left[\mathrm{KL}(P_\theta^{adv}\|P_{\theta_0}^{adv}) + \mathrm{KL}(P_\theta^{adv}\|P_\theta^{clean})\right]
\end{equation}
where $\mathrm{KL}(\cdot\|\cdot)$ denotes KL divergence. 
To sum up, with additional regularisation, the weights of the vision encoder $f_\theta$ are updated as follows:
\begin{equation}\label{eq:final_loss}
\begin{split}
    \theta' = \arg\min_\theta \{ \:
        \mathcal{L}_{CLIP}& \left(
        \{(x_i+\delta_i,t_i)\}_{i=1}^N
        \,\boldsymbol{\vert}\, \theta,\phi
        \right)\\
        &+ 
        \mathcal{L}_{logit} + 
        \mathcal{L}_{feat} \:
    \}
\end{split}
\end{equation}
We illustrate in \cref{fig:AdvFLYP_reg} the outer minimisation process where the contrastive objective 
$\mathcal{L}_{CLIP}\left(\{(x_i+\delta_i,t_i)\}_{i=1}^N
        \,\boldsymbol{\vert}\, \theta,\phi\right)$
and the regularisation terms are formulated to update the vision encoder $f_\theta$.
Our AdvFLYP can be practically interpreted as resuming the training of the pretrained CLIP, except that the text encoder $g_\phi$ is kept frozen and that adversarial images, rather than natural images, are aligned with their corresponding texts via a contrastive loss.
As opposed to prior AFT methods, this paradigm is simple and intuitive, respecting the pretraining pattern of CLIP.
\cref{alg:AdvFLYP} summarises the AdvFLYP paradigm.

\begin{algorithm}[t]
\caption{PyTorch-style pseudocode for AdvFLYP} 
\label{alg:AdvFLYP}
\vspace{-1.ex}
\begin{lstlisting}[style=Pytorch,escapeinside={(@}{@)}]
# f:      target vision encoder
# F:      frozen original vision encoder
# g:      frozen text encoder
# L_clip: contrastive objective function
# D:      collected webscale image-text pairs
for (X, T) in D:
    # generate adversarial perturbations
    delta = PGD(f, g, (X,T), L_clip)
    # contrastive loss
    l_clip = L_clip(f, g, (X+delta,T))
    l_final = l_clip
    # obtain normalised embeddings
    X=f(X+delta),X_c=f(X),X_ori=F(X+delta),T=g(T)
    # feature-level regularization
    l_feat=(X-X_c).norm(-1)+(X-X_ori).norm(-1)
    # compute probability logit
    P=(X@T.t()).softmax(-1)
    P_c=(X_c@T.t()).softmax(-1)
    P_ori=(X_ori@T.t()).softmax(-1)
    # logit-level regularization
    l_logit=P*(P/P_c).log()+P*(P/P_ori).log()
    # regularised AdvFLYP
    if regularised:
      l_final += (l_logit.mean()+l_feat.mean())
    # update theta w.r.t. final loss
    l_final.backward()
    optimizer.step()
return f
\end{lstlisting}
\vspace{-1.ex}
\end{algorithm}
\section{Experiments}
\label{sec:experiment}
In this section, we conduct extensive experiments to evaluate the adversarial robustness and the retention of zero-shot capabilities of AdvFLYP, compared to prior AFT methods.

\subsection{Baselines and Datasets}
\noindent
\textbf{Baselines.}
Based on their released code, we re-implement prior AFT methods, where a proxy dataset with labelled classes is involved to create adversarial images for adversarial training.
TeCoA \cite{mao2023understanding} is the fundamental paradigm that finetunes the vision encoder $f_\theta$ to correctly classify the adversarial images by minimising the cross-entropy \wrt the ground-truth labels (see \cref{eq:tecoa_outer_loop}).
Other baselines are based on this paradigm.
PMG-AFT \cite{wang2024pre} leverages the original CLIP to regularise the finetuning process by introducing KL-divergence losses on the logit level of the target model.
TGA-ZSR \cite{yu2024text} introduces text-guided attention, which employs the frozen textual features of the correct class labels to signal the importance of each image area, and proposes a loss term that aligns the attention maps with the ones induced by the original CLIP.
In addition to these supervised methods, we further implement an unsupervised AFT method FARE \cite{schlarmann2024robust}, which dispenses with the need for class labels. 
The adversarial images are created such that the $L_2$ distance between their embeddings and the clean counterparts is maximised, and the vision encoder $f_\theta$ is updated to minimise this distance.
Following TeCoA, we employ the training set of ImageNet \cite{deng2009imagenet} for implementing these baselines, which contains more than 1.2M labelled images in the training set.

\noindent
\textbf{Datasets.} We evaluate all methods on 14 downstream classification datasets spanning various domains, which include
general object recognition datasets 
CIFAR10 \citep{krizhevsky2009learning}, 
CIFAR100 \citep{krizhevsky2009learning}, 
STL10 \citep{coates2011analysis}, 
Caltech101 \citep{fei2006one} and 
Caltech256 \citep{griffin2007caltech};
fine-grained recognition datasets 
OxfordPets \citep{parkhi2012cats}, 
Flowers102 \citep{nilsback2008automated}, 
Food101 \citep{bossard2014food},
StanfordCars \citep{krause20133d}; 
scene recognition datasets 
SUN397 \citep{xiao2010sun} and
Country211 \citep{radford2021learning};
domain-specific datasets 
FGVCAircraft \citep{maji2013fine},
EuroSAT \citep{helber2019eurosat}, and
DTD \citep{cimpoi2014describing}.
Since the baselines are finetuned on ImageNet, we additionally employ four cross-domain variants of ImageNet to evaluate transferability of robustness to other domains, including ImageNet-R \cite{hendrycks2021many}, ImageNet-A \cite{hendrycks2021natural}, ImageNet-S \cite{wang2019learning}, and ObjectNet \cite{NEURIPS2019_97af07a1}. We provide more information on these datasets in Appendix \cref{App_sec:dataset_info}.

\subsection{Implementation Details}
Following prior AFT methods, we implement all experiments on the pretrained CLIP with the ViT-B/32 vision encoder. 
The batch size is set to 256. The initial learning rate is $1e-4$, which is adjusted with a cosine annealing scheduler.
During AFT, we employ PGD-2 \cite{carlini2017towards} to create adversarial images, 
which updates the batchwise adversarial perturbations in two iterative steps,
with both the attack strength and the step size set to $1/255$.
We employ a toy classification dataset TinyImageNet \cite{le2015tiny} to evaluate the robustness of the model at the end of each epoch, and terminate the process when there is no improvement for 10 epochs.
At test time, the finetuned model is deployed for downstream classification datasets, where the the adversarial images are created by maximising the cross-entropy of the images \wrt the true labels, irrespective of the training objective during AFT.
All experiments are conducted on a single NVIDIA A100-SXM-64GB GPU. 

\subsection{Results and Discussions}

\begin{table*}[t]
    \centering
    \resizebox{\textwidth}{!}{
    \begin{tabular}{c|c|cccccccccccccc|c}
    \toprule
    \multicolumn{2}{c|}{\raisebox{3\height}{(\%)}} 
    & \spin{CIFAR10} & \spin{CIFAR100} & \spin{STL10} & \spin{Caltech101} & \spin{Caltech256} & \spin{OxfordPets} & \spin{Flowers102} & \spin{Food101} & \spin{StanfordCars} & \spin{SUN397} & \spin{Country211} & \spin{FGVCAircraft} & \spin{EuroSAT} & \spin{DTD} & \spin{avg.}\\
    \midrule\midrule
    \multirow{7}{*}{\spin{PGD}}
    & \color{lightgray}CLIP 
    & \color{lightgray}0.77 & \color{lightgray}0.18 &\color{lightgray}11.33 &\color{lightgray}14.87 &\color{lightgray}8.58 &\color{lightgray}1.04 &\color{lightgray}1.04 &\color{lightgray}0.67 &\color{lightgray}0.03 &\color{lightgray}1.14 &\color{lightgray}0.03 &\color{lightgray}0.00 &\color{lightgray}0.03 &\color{lightgray}3.19 &\color{lightgray}3.06 \\
    & FARE 
    &21.44 &10.05 &59.82 &54.51 &44.73 &37.18 &16.21 &15.90 &6.47 &15.05 &0.74 &1.65 &6.00 &17.13 &21.92  \\
    & TeCoA 
    &37.54 &19.87 &75.59 &69.02 &59.41 &61.16 &31.05 &27.89 &13.83 &31.07 &3.22 &5.40 &\bf15.37 &22.39 &33.77 \\
    & PMG-AFT 
    &42.06 &21.71 &77.34 &71.91 &60.88 &\underline{63.83} &33.81 &32.84 &17.83 &32.19 &3.25 &6.09 &\un{14.71} &23.40 &35.85 \\
    & TGA-ZSR 
    &37.73 &19.01 &\underline{78.89} &\bf75.64 &\bf65.13 &\bf73.34 &34.41 &\underline{37.27} &19.29 &\bf37.38 &3.42 &\bf8.97 &14.36 &\un{26.17} &\un{37.93} \\
    \cmidrule{2-17}
    & 
    AdvFLYP 
    &\underline{45.55} &\underline{24.36} &75.15 &70.81 &60.35 &56.77 &\underline{35.13} &34.46 &\un{24.32} &34.28 &\un{3.82} &7.02 &5.76 &26.06 &35.99 \\
    & 
    $\mathrm{AdvFLYP}_{full}$ 
    &\bf52.24 &\bf27.89 &\bf78.92 &\underline{73.15} &\underline{62.22} &61.16 &\bf39.50 &\bf39.18 &\bf26.86 &\un{36.34} &\bf4.14 &\un{7.17} &12.88 &\bf26.44 &\bf39.15 \\
    
    \midrule\midrule
    
    \multirow{7}{*}{\spin{CW}}
    & \color{lightgray}CLIP 
    & \color{lightgray}0.72 & \color{lightgray}0.15 &\color{lightgray}12.26 &\color{lightgray}21.29 &\color{lightgray}9.69 &\color{lightgray}1.69 &\color{lightgray}1.42 &\color{lightgray}1.12 &\color{lightgray}2.36 &\color{lightgray}1.73 &\color{lightgray}0.08 &\color{lightgray}0.00 &\color{lightgray}0.05 &\color{lightgray}2.55 &\color{lightgray}3.94 \\
    & FARE 
    &22.11 &10.44 &61.16 &59.66 &46.65 &39.49 &17.68 &18.98 &9.90 &16.76 &0.95 &1.68 &5.67 &15.80 &23.35  \\
    & TeCoA 
    &36.46 &18.68 &75.40 &70.14 &59.04 &60.97 &29.58 &26.45 &13.70 &29.95 &2.71 &4.68 &\un{13.83} &21.01 &33.04  \\
    & PMG-AFT 
    &40.53 &20.28 &77.21 &72.92 &60.20 &\un{63.67} &31.47 &31.45 &16.65 &30.54 &2.80 &5.04 &13.59 &21.86 &34.87  \\
    & TGA-ZSR 
    &39.50 &19.83 &\bf79.53 &\bf77.16 &\bf65.18 &\bf72.61 &33.89 &\un{38.42} &19.96 &\bf37.81 &\bf3.74 &\bf8.55 &\bf14.06 &\bf25.16 &\un{38.24}  \\
    \cmidrule{2-17}
    & 
    AdvFLYP 
    &\un{45.13} &\un{24.06} &75.21 &72.10 &60.41 &57.32 &\un{34.95} &34.49 &\un{24.69} &34.07 &3.39 &\un{6.51} &5.29 &\un{24.95} &35.90  \\
    & 
    $\mathrm{AdvFLYP}_{full}$ 
    &\bf50.51 &\bf26.21 &\un{78.64} &\un{73.65} &\un{61.75} &60.94 &\bf38.06 &\bf38.91 &\bf25.80 &\un{35.18} &\un{3.58} &6.12 &12.47 &24.36 &\bf38.30  \\
    
    \midrule\midrule
    
    \multirow{7}{*}{\spin{AutoAttack}}
    & \color{lightgray}CLIP 
    & \color{lightgray}0.00 & \color{lightgray}0.06 &\color{lightgray}0.00 &\color{lightgray}0.43 &\color{lightgray}0.10 &\color{lightgray}0.00 &\color{lightgray}0.02 &\color{lightgray}0.00 &\color{lightgray}0.00 &\color{lightgray}0.01 &\color{lightgray}0.00 &\color{lightgray}0.03 &\color{lightgray}0.08 &\color{lightgray}0.05 &\color{lightgray}0.05 \\
    & FARE 
    &20.01 &9.02 &58.89 &53.53 &43.65 &36.39 &14.98 &15.13 &5.93 &13.55 &0.58 &1.08 &5.55 &15.27 &20.97  \\
    & TeCoA 
    &35.39 &17.95 &74.86 &68.34 &58.21 &\un{59.96} &28.90 &25.25 &11.83 &28.63 &2.60 &4.50 &\bf13.11 &20.85 &32.17  \\
    & PMG-AFT 
    &39.23 &19.40 &\un{76.79} &\un{71.14} &59.34 &\bf62.42 &30.69 &29.86 &14.48 &29.26 &2.59 &4.65 &\un{12.83} &21.76 &33.89  \\
    & TGA-ZSR 
    &0.01 &0.00 &0.11 &0.29 &0.11 &0.06 &0.00 &0.01 &0.00 &0.01 &0.03 &0.06 &0.10 &0.16 &0.07  \\
    \cmidrule{2-17}
    & 
    AdvFLYP 
    &\un{44.37} &\un{23.01} &74.88 &70.31 &\un{59.70} &55.96 &\un{34.22} &\un{32.99} &\un{22.47} &\un{32.94} &\un{3.20} &\un{5.70} &4.67 &\un{24.73} &\un{34.94}  \\
    & 
    $\mathrm{AdvFLYP}_{full}$ 
    &\bf50.15 &\bf25.99 &\bf78.50 &\bf72.47 &\bf61.31 &59.88 &\bf37.81 &\bf37.53 &\bf24.61 &\bf34.44 &\bf3.46 &\bf5.73 &11.46 &\bf24.95 &\bf37.74  \\
    
    \midrule\midrule
    
    \multirow{7}{*}{\spin{$\overline{\mathrm{AVG}}$}}
    & \color{lightgray}CLIP 
    & \color{lightgray}0.50 & \color{lightgray}0.13 &\color{lightgray}7.86 &\color{lightgray}12.19 &\color{lightgray}6.12 &\color{lightgray}0.91 &\color{lightgray}0.82 &\color{lightgray}0.60 &\color{lightgray}0.80 &\color{lightgray}0.96 &\color{lightgray}0.04 &\color{lightgray}0.01 &\color{lightgray}0.06 &\color{lightgray}1.93 &\color{lightgray}2.35 \\
    & FARE 
    &21.19 &9.84 &59.96 &55.90 &45.01 &37.68 &16.29 &16.67 &7.43 &15.12 &0.75 &1.47 &5.74 &16.06 &22.08  \\
    & TeCoA 
    &36.46 &18.83 &75.28 &69.17 &58.89 &\un{60.70} &29.84 &26.53 &13.12 &29.88 &2.85 &4.86 &\bf14.10 &21.42 &33.00  \\
    & PMG-AFT 
    &40.61 &20.46 &\un{77.11} &\un{71.99} &60.14 &\bf63.31 &31.99 &31.38 &16.32 &30.66 &2.88 &5.26 &\un{13.71} &\un{22.34} &34.87  \\
    & TGA-ZSR 
    &25.75 &12.95 &52.84 &51.03 &43.47 &48.67 &22.77 &25.23 &13.08 &25.07 &2.40 &5.86 &9.51 &17.16 &25.41  \\
    \cmidrule{2-17}
    & 
    AdvFLYP 
    &\un{45.02} &\un{23.81} &75.08 &71.07 &\un{60.15} &56.68 &\un{34.76} &\un{33.98} &\un{23.83} &\un{33.76} &\un{3.47} &\bf6.41 &5.24 &\bf25.25 &\un{35.61}  \\
    &
    $\mathrm{AdvFLYP}_{full}$ 
    &\bf50.97 &\bf26.70 &\bf78.69 &\bf73.09 &\bf61.76 &60.66 &\bf38.46 &\bf38.54 &\bf25.76 &\bf35.32 &\bf3.73 &\un{6.34} &12.27 &\bf25.25 &\bf38.39  \\
    \bottomrule 
    \end{tabular}}
    \vspace{-1em}
    \caption{Classification accuracy (\%) 
    on 14 downstream datasets tested with three adversarial attack algorithms.
    We highlight the \textbf{best} and \underline{second best} result.
    }
    \label{tab:test_eps1}
\end{table*}

\begin{table*}[t]
    \centering
    \resizebox{\textwidth}{!}{
    \begin{tabular}{c|cccccccccccccc|c}
    \toprule
    \raisebox{3\height}{(\%)} 
    & \spin{CIFAR10} & \spin{CIFAR100} & \spin{STL10} & \spin{Caltech101} & \spin{Caltech256} & \spin{OxfordPets} & \spin{Flowers102} & \spin{Food101} & \spin{StanfordCars} & \spin{SUN397} & \spin{Country211} & \spin{FGVCAircraft} & \spin{EuroSAT} & \spin{DTD} & \spin{avg.}\\
    \midrule
    \color{lightgray}CLIP 
    & \color{lightgray}85.05 & \color{lightgray}57.18 &\color{lightgray}96.41 &\color{lightgray}86.19 &\color{lightgray}82.04
    &\color{lightgray}87.27 &\color{lightgray}65.62 &\color{lightgray}83.83 &\color{lightgray}52.13 &\color{lightgray}58.87 &\color{lightgray}15.26 &\color{lightgray}20.16 &\color{lightgray}38.32 &\color{lightgray}40.11 &\color{lightgray}62.03 \\
    FARE 
    &\bf80.49 &\bf55.21 &\bf95.14 &\bf86.42 &\bf81.97 &\bf87.46 &\bf61.94 &\bf76.24 &\bf48.68 &\bf58.75 &\bf12.46 &\bf18.69 &\un{28.74} &\bf40.11 &\bf59.45 \\
    TeCoA 
    &70.60 &40.11 &91.48 &81.31 &76.76 &82.01 &51.63 &55.79 &34.12 &53.29 &8.14 &13.44 &26.97 &33.30 &51.35 \\
    PMG-AFT 
    &75.36 &43.94 &\un{92.91} &\un{84.50} &\un{78.43} &\un{84.11} &55.23 &64.59 &42.07 &55.31 &9.23 &15.09 &24.36 &34.31 &54.25 \\
    TGA-ZSR 
    &\un{77.22} &41.77 &91.60 &78.46 &78.09 &81.36 &52.92 &65.74 &37.81 &52.87 &9.86 &12.27 &\bf30.03 &35.96 &53.28 \\
    \midrule
    AdvFLYP 
    &74.15 &\un{48.86} &90.70 &83.82 &76.98 &79.69 &52.50 &64.72 &44.98 &55.58 &10.54 &16.08 &23.79 &\un{37.34} &54.27 \\
    $\mathrm{AdvFLYP}_{full}$ 
    &75.76 &46.91 &92.09 &84.40 &77.44 &83.89 &\un{57.67} &\un{69.48} &\un{48.10} &\un{57.26} &\un{11.00} &\un{16.80} &24.43 &36.54 &
    \un{55.84} \\
    \bottomrule 
    \end{tabular}
    }
    \vspace{-1em}
    \caption{Accuracy (\%) on clean images of 14 downstream datasets.}
    \label{tab:test_clean}
    \vspace{-.9em}
\end{table*}

\noindent
\textbf{Adversarial robustness at $\epsilon=1/255$.}
\cref{tab:test_eps1} reports the accuracy of all baselines under three attack methods, PGD-10 \cite{madry2018towards}, CW-10 \cite{carlini2017towards}, and AutoAttack \cite{croce2020reliable}. We also report the average accuracy over these three attacks for a comprehensive assessment. 
From \cref{tab:test_eps1}, it can be seen 
that when following the basic pretraining pattern of CLIP in AFT, \ie, updating the model weights by matching adversarial web images with their corresponding texts, achieves a robust accuracy of $35.61\%$ averaged over all datasets and attack methods.
The basic AdvFLYP paradigm outperforms TeCoA and their subsequent advancements with regularisation, which leverage a well-curated dataset with labelled classes to perform AFT with a slightly larger amount of training data.
This challenges the current \textit{de facto} standard practice of finetuning the CLIP model on a labelled class dataset via a cross-entropy loss,
which is consistent with the evaluation process on downstream tasks but deviates from the pretraining behaviour.
With additional regularisation on the feature and logit levels of the adversarial images during finetuning, the zero-shot robustness of 
$\textrm{AdvFLYP}_{full}$
is further enhanced, with an average robust accuracy of $38.39\%$, showing the effectiveness of the regularisation terms.
\cref{tab:test_clean} reports the clean accuracy of all baselines. The unsupervised AFT method FARE \cite{schlarmann2024robust} achieves the best clean accuracy. 
Among the supervised baselines, AdvFLYP fares the best, and 
$\textrm{AdvFLYP}_{full}$
further mitigates the trade-off of clean accuracy, with an average accuracy of $55.84\%$.
To sum up, despite its simplicity, AdvFLYP significantly outperforms TeCoA, which is the standard AFT paradigm in prior methods, highlighting the importance of considering CLIP's pretraining pattern when performing AFT.
Additional logit- and feature-level regularisation proves effective in improving the
generalisation of the finetuned model, significantly surpassing the regularisation-based advancements of TeCoA.

\noindent
\textbf{Adversarial robustness at stronger attacks.} \cref{tab:stronger_attacks} reports the robustness under three attack methods with a higher attack budget of $\epsilon=2/255$ and $\epsilon=4/255$. 
The average results over 14 downstream datasets show that AdvFLYP and its $\textrm{AdvFLYP}_{full}$ consistently outperform previous AFT baselines.
We provide the full tables in Appendix \cref{App_sec:more_experiment_results}.


\begin{table}[t]
    \centering
    \resizebox{\linewidth}{!}{
    \begin{tabular}{c|c|ccc|c}
    \toprule
    \multicolumn{2}{c|}{(\%)}& PGD & CW & AA & $\overline{\textrm{AVG}}$ \\
    \midrule
    
    \multirow{4}{*}{$\epsilon=2/255$} 
    & TeCoA & 17.89 & 18.67 & 16.57 & 17.71 \\
    & PMG-AFT & 18.66 & 19.25 & 17.00 & 18.31\\
    \cmidrule{2-6}
    &
    AdvFLYP & \un{19.94} & \un{21.27} & \un{19.01} & \un{20.07}\\
    &
    $\textrm{AdvFLYP}_{full}$ & \bf22.02 & \bf22.69 & \bf20.35 & \bf21.69 \\
    \midrule
    \midrule
    \multirow{4}{*}{$\epsilon=4/255$} 
    & TeCoA & 4.40 & 5.37 & 2.49 & 4.09\\
    & PMG-AFT & 4.44 & 5.30 & 2.26 & 4.00\\
    \cmidrule{2-6}
    &
    AdvFLYP & \un{6.18} & \un{7.27} & \bf4.15 & \un{5.87}\\
    &
    $\textrm{AdvFLYP}_{full}$ & \bf6.45 & \bf7.36 & \un{3.98} & \bf5.93\\
    \bottomrule
    \end{tabular}
    }
    \vspace{-1em}
    \caption{Adversarial robustness under stronger attack budget $\epsilon$. The reported values are accuracy averaged over 14 datasets.}
    \label{tab:stronger_attacks}
    \vspace{-.7em}
\end{table}

\subsection{Ablation on Regularisation}\label{sec:ablation_regularisation}
\citet{wang2024pre} show that imposing CLIP-guided regularisation on the logit level strengthens generalisation on both adversarial and clean examples.
Our implementations confirm this finding, as PMG-AFT \cite{wang2024pre} outperforms TeCoA \cite{mao2023understanding} in terms of robust and clean accuracy (\cref{tab:test_eps1}).
\citet{wang2024pre} also show that feature-level regularisation does not lead to robustness or generalisation gains.
In this work, we find that our AdvFLYP paradigm exhibits different behaviour, with logit- and feature-level regularisation benefiting robustness and clean accuracy, respectively.
\cref{tab:regularisation_ablation} reports our ablative studies. 
It can be seen that the contrastive objective between the adversarial images and texts is central to the robustness gains (\textit{a}).
Regularising AdvFLYP on the logit level markedly improve downstream robustness, but at a slight cost of clean accuracy (comparing \textit{a} and \textit{b}). 
In contrast, regularising AdvFLYP only on the feature level plays an important role in retaining zero-shot abilities of CLIP, as evidenced by the markedly improved clean accuracy (\textit{c}). 
This behaviour is in stark contrast with the findings of PMG-AFT \cite{wang2024pre}
, where logit-level regularisation is shown to improve both robustness and clean accuracy, while feature-level regularisation does not benefit either.
This is due to the fact that AdvFLYP leverages image-text pairs from the web, which are highly diverse and noisy. 
Producing the adversarial images of these web images by maximising the contrastive loss results in distorted vision embeddings, which can compromise the vision encoder when used for finetuning.
Therefore, penalising the vision encoder for producing image embeddings that shift drastically in the normalised embedding space is effective in preserving the zero-shot abilities.
In this work, we incorporate logit- and feature-level regularisers in our AdvFLYP to achieve a sweet spot of robustness and clean accuracy without further tuning their weights.
We provide more results and analyses as well as more ablations regarding other training settings of AdvFLYP in Appendix \cref{App_sec:more_ablations}.

\begin{table}[t]
    \centering
    \resizebox{\linewidth}{!}{
    \begin{tabular}{c|c|cc|c}
    \toprule
    \multicolumn{2}{c|}{(\%)} & \begin{tabular}{@{}c@{}}AutoAttack \\ ($\epsilon=1/255$)\end{tabular} & Clean & Avg.\\
    \midrule
    & \textcolor{lightgray}{CLIP} & \textcolor{lightgray}{0.05} & \textcolor{lightgray}{62.03} & 
    \textcolor{lightgray}{31.04} \\
    \cmidrule{2-5}
    (a) & AdvFLYP & 34.94 & 54.27 & 44.61\\
    (b) & AdvFLYP + $\mathcal{L}_{logit}$& \un{37.40} & 53.86 & 45.63 \\
    (c) & AdvFLYP + $\mathcal{L}_{feat}$& 35.01 & \bf57.50 & \un{46.26}\\
    (d) & AdvFLYP + $\mathcal{L}_{logit}$ + $\mathcal{L}_{feat}$& \bf37.74 & \un{55.84} & \bf46.79\\
    \bottomrule
    \end{tabular}
    }
    \vspace{-1em}
    \caption{Adversarial robustness and clean accuracy of baselines adopting different combinations of finetuning objectives. Variant (d) amounts to $\textrm{AdvFLYP}_{full}$.}\label{tab:regularisation_ablation}
    \vspace{-1.1em}
\end{table}

\subsection{More Discussion on Data}
The aim of this work is to rethink the current \textit{de facto} standard AFT paradigm, which caters to the downstream classification tasks by finetuning $f_\theta$ on a well-curated classification dataset.
We show that by respecting the pretraining pattern of VLMs, AdvFLYP conducting AFT based on noisy web data still outperforms previous methods that leverage a well-curated proxy dataset.
To further investigate the robustness gains of prior AFT methods, we evaluate PMG-AFT on several popular ImageNet variants, which share the pre-defined classes but have distinctly different distributions.
As shown in \cref{tab:in_variants}, PMG-AFT achieves significantly improved robustness on ImageNet.
However, this robustness gain is limited on other data domains, as evidenced by the smaller improvement on other variant datasets.
It is also noteworthy that PMG-AFT even fares better than the original CLIP on the clean images of ImageNet, which implies that the model has memorised the data distribution of ImageNet in some way, even though it is finetuned on adversarial examples.
To elucidate the contributions of training data, we employ the pretrained generative VLM Qwen2.5-VL-3B-Instruct
to caption the training set of ImageNet, and perform contrastive finetuning on the captioned dataset. 
Experiments show that enriching the descriptive texts of ImageNet and finetuning the model via a contrastive objective improves robustness over current AFT methods, but to a lesser extent compared to AdvFLYP.
We include the results in Appendix \cref{App_sec:training_data}.

\begin{table}[t]
    \centering
    \resizebox{\linewidth}{!}{
    \begin{tabular}{c|c|c|cccc|c}
    \toprule
    \multicolumn{2}{c|}{(\%)}& IN & -R & -A & -S & Net & Avg. \\
    \midrule
    \multirow{3}{*}{\spin{Adv.}}& \color{lightgray}CLIP  &\color{lightgray}1.14  &\color{lightgray}6.32  &\color{lightgray}0.09  &\color{lightgray}5.08  &\color{lightgray}0.12  &\color{lightgray}2.91 \\
    & PMG-AFT  &\textbf{39.42} &40.95  &3.01  &25.07  &9.61  &19.66 \\
    &
    $\textrm{AdvFLYP}_{full}$ &31.15  &\textbf{42.45}  &\textbf{4.04}  &\textbf{25.36}  &\textbf{10.72}  &\textbf{20.64} \\
    \midrule
    \multirow{3}{*}{\spin{Clean}}& \color{lightgray}CLIP  &\color{lightgray}59.75  &\color{lightgray}65.10  &\color{lightgray}29.71  &\color{lightgray}39.47   &\color{lightgray}33.28  &\color{lightgray}41.89 \\
    & PMG-AFT  &\textbf{61.74} &59.94  &12.32  &36.91  &21.47  &32.66 \\
    &
    $\textrm{AdvFLYP}_{full}$ &52.43 &\textbf{60.40}  &\textbf{13.32}  &\textbf{37.07}  &\textbf{22.36}  &\textbf{33.29}  \\
    \bottomrule
    \end{tabular}}
    \vspace{-1em}
    \caption{Comparison of PMG-AFT finetuned on ImageNet (IN) and AdvFLYP on web image-text data. The adversarial accuracy is tested under PGD-10 with $\epsilon=1/255$.}
    \label{tab:in_variants}
    \vspace{-1.1em}
\end{table}

\subsection{AdvFLYP \textit{versus} FLYP}
Although we draw inspiration from FLYP \cite{goyal2023finetune}, AdvFLYP differs fundamentally from FLYP in both motivations and implementations.
While FLYP aims to improve generalisation of the model on OOD data, AdvFLYP seeks to strengthen the zero-shot adversarial robustness of VLMs.
In terms of implementations, although FLYP employs the contrastive loss, it still finetunes the model on ID data with labelled classes and a fixed textual template, ignoring the fact that classes may overlap within the same training batch.
In contrast, AdvFLYP performs real contrastive finetuning on image-text pairs, showing that respecting CLIP's pretraining data distribution and training objective yields more transferable robustness gains and mitigates the trade-off of clean accuracy.
We experiment FLYP na\"{i}vely on ImageNet in the context of AFT and find that it does not lead to better results compared to prior AFT methods. The results are included in \cref{sec:naive_flyp}.

\section{Conclusion}\label{sec:conclusion}
Current AFT methods for CLIP invariably leverages a proxy dataset with labelled classes, where the vision encoder is finetuned via a cross-entropy loss to cater to downstream classification tasks.
We rethink this paradigm and believe that it largely overlooks the important roles of training data distributions and objectives, which drastically deviates from the pretraining behaviour of CLIP, causing limited robustness gains and reduced zero-shot knowledge.
We propose a simple yet effective AdvFLYP paradigm, which respects CLIP's pretraining behaviour in AFT rather than catering to downstream tasks.
Additionally, we find that logit- and feature-level regularisation on top of AdvFLYP benefit robustness and clean accuracy, respectively.
Experiments on 14 datasets show that AdvFLYP outperforms the current \textit{de facto} standard AFT paradigm consistently under various attack scenarios.
We hope this work raises to the community the importance of considering the pretraining of VLMs and establish a new standard in adversarial finetuning.

\clearpage
\section*{Acknowledgments}
The authors acknowledge the CINECA award under the ISCRA initiative for the availability of high-performance computing resources and support.
This work was supported by Mobility Program of the European Lighthouse of AI for Sustainability (ELIAS) under Grant Agreement No. 101120237.

{
    \small
    \bibliographystyle{ieeenat_fullname}
    \bibliography{main}

\begin{thebibliography}{66}
\providecommand{\natexlab}[1]{#1}
\providecommand{\url}[1]{\texttt{#1}}
\expandafter\ifx\csname urlstyle\endcsname\relax
  \providecommand{\doi}[1]{doi: #1}\else
  \providecommand{\doi}{doi: \begingroup \urlstyle{rm}\Url}\fi

\bibitem[Barbu et~al.(2019)Barbu, Mayo, Alverio, Luo, Wang, Gutfreund, Tenenbaum, and Katz]{NEURIPS2019_97af07a1}
Andrei Barbu, David Mayo, Julian Alverio, William Luo, Christopher Wang, Dan Gutfreund, Josh Tenenbaum, and Boris Katz.
\newblock Objectnet: A large-scale bias-controlled dataset for pushing the limits of object recognition models.
\newblock In \emph{Advances in Neural Information Processing Systems}. Curran Associates, Inc., 2019.

\bibitem[Bossard et~al.(2014)Bossard, Guillaumin, and Van~Gool]{bossard2014food}
Lukas Bossard, Matthieu Guillaumin, and Luc Van~Gool.
\newblock Food-101--mining discriminative components with random forests.
\newblock In \emph{European conference on computer vision}, pages 446--461. Springer, 2014.

\bibitem[Carlini and Wagner(2017)]{carlini2017towards}
Nicholas Carlini and David Wagner.
\newblock Towards evaluating the robustness of neural networks.
\newblock In \emph{2017 ieee symposium on security and privacy (sp)}, pages 39--57. Ieee, 2017.

\bibitem[Cimpoi et~al.(2014)Cimpoi, Maji, Kokkinos, Mohamed, and Vedaldi]{cimpoi2014describing}
Mircea Cimpoi, Subhransu Maji, Iasonas Kokkinos, Sammy Mohamed, and Andrea Vedaldi.
\newblock Describing textures in the wild.
\newblock In \emph{Proceedings of the IEEE conference on computer vision and pattern recognition}, pages 3606--3613, 2014.

\bibitem[Coates et~al.(2011)Coates, Ng, and Lee]{coates2011analysis}
Adam Coates, Andrew Ng, and Honglak Lee.
\newblock An analysis of single-layer networks in unsupervised feature learning.
\newblock In \emph{Proceedings of the fourteenth international conference on artificial intelligence and statistics}, pages 215--223. JMLR Workshop and Conference Proceedings, 2011.

\bibitem[Croce and Hein(2020)]{croce2020reliable}
Francesco Croce and Matthias Hein.
\newblock Reliable evaluation of adversarial robustness with an ensemble of diverse parameter-free attacks.
\newblock In \emph{International conference on machine learning}, pages 2206--2216. PMLR, 2020.

\bibitem[Deng et~al.(2009)Deng, Dong, Socher, Li, Li, and Fei-Fei]{deng2009imagenet}
Jia Deng, Wei Dong, Richard Socher, Li-Jia Li, Kai Li, and Li Fei-Fei.
\newblock Imagenet: A large-scale hierarchical image database.
\newblock In \emph{2009 IEEE conference on computer vision and pattern recognition}, pages 248--255. Ieee, 2009.

\bibitem[Dong et~al.(2025)Dong, Koniusz, Zhang, Zhu, Liu, Qu, and Ong]{dong2025improving}
Junhao Dong, Piotr Koniusz, Yifei Zhang, Hao Zhu, Weiming Liu, Xinghua Qu, and Yew-Soon Ong.
\newblock Improving zero-shot adversarial robustness in vision-language models by closed-form alignment of adversarial path simplices.
\newblock In \emph{Forty-second International Conference on Machine Learning}, 2025.

\bibitem[Fei-Fei et~al.(2006)Fei-Fei, Fergus, and Perona]{fei2006one}
Li Fei-Fei, Robert Fergus, and Pietro Perona.
\newblock One-shot learning of object categories.
\newblock \emph{IEEE transactions on pattern analysis and machine intelligence}, 28\penalty0 (4):\penalty0 594--611, 2006.

\bibitem[Goyal et~al.(2023)Goyal, Kumar, Garg, Kolter, and Raghunathan]{goyal2023finetune}
Sachin Goyal, Ananya Kumar, Sankalp Garg, Zico Kolter, and Aditi Raghunathan.
\newblock Finetune like you pretrain: Improved finetuning of zero-shot vision models.
\newblock In \emph{Proceedings of the IEEE/CVF Conference on Computer Vision and Pattern Recognition}, pages 19338--19347, 2023.

\bibitem[Griffin et~al.(2007)Griffin, Holub, Perona, et~al.]{griffin2007caltech}
Gregory Griffin, Alex Holub, Pietro Perona, et~al.
\newblock Caltech-256 object category dataset.
\newblock Technical report, Technical Report 7694, California Institute of Technology Pasadena, 2007.

\bibitem[He et~al.(2016)He, Zhang, Ren, and Sun]{he2016deep}
Kaiming He, Xiangyu Zhang, Shaoqing Ren, and Jian Sun.
\newblock Deep residual learning for image recognition.
\newblock In \emph{Proceedings of the IEEE conference on computer vision and pattern recognition}, pages 770--778, 2016.

\bibitem[Helber et~al.(2019)Helber, Bischke, Dengel, and Borth]{helber2019eurosat}
Patrick Helber, Benjamin Bischke, Andreas Dengel, and Damian Borth.
\newblock Eurosat: A novel dataset and deep learning benchmark for land use and land cover classification.
\newblock \emph{IEEE Journal of Selected Topics in Applied Earth Observations and Remote Sensing}, 12\penalty0 (7):\penalty0 2217--2226, 2019.

\bibitem[Hendrycks et~al.(2021{\natexlab{a}})Hendrycks, Basart, Mu, Kadavath, Wang, Dorundo, Desai, Zhu, Parajuli, Guo, et~al.]{hendrycks2021many}
Dan Hendrycks, Steven Basart, Norman Mu, Saurav Kadavath, Frank Wang, Evan Dorundo, Rahul Desai, Tyler Zhu, Samyak Parajuli, Mike Guo, et~al.
\newblock The many faces of robustness: A critical analysis of out-of-distribution generalization.
\newblock In \emph{Proceedings of the IEEE/CVF international conference on computer vision}, pages 8340--8349, 2021{\natexlab{a}}.

\bibitem[Hendrycks et~al.(2021{\natexlab{b}})Hendrycks, Zhao, Basart, Steinhardt, and Song]{hendrycks2021natural}
Dan Hendrycks, Kevin Zhao, Steven Basart, Jacob Steinhardt, and Dawn Song.
\newblock Natural adversarial examples.
\newblock In \emph{Proceedings of the IEEE/CVF conference on computer vision and pattern recognition}, pages 15262--15271, 2021{\natexlab{b}}.

\bibitem[Jia et~al.(2021)Jia, Yang, Xia, Chen, Parekh, Pham, Le, Sung, Li, and Duerig]{jia2021scaling}
Chao Jia, Yinfei Yang, Ye Xia, Yi-Ting Chen, Zarana Parekh, Hieu Pham, Quoc Le, Yun-Hsuan Sung, Zhen Li, and Tom Duerig.
\newblock Scaling up visual and vision-language representation learning with noisy text supervision.
\newblock In \emph{International conference on machine learning}, pages 4904--4916. PMLR, 2021.

\bibitem[Krause et~al.(2013)Krause, Stark, Deng, and Fei-Fei]{krause20133d}
Jonathan Krause, Michael Stark, Jia Deng, and Li Fei-Fei.
\newblock 3d object representations for fine-grained categorization.
\newblock In \emph{Proceedings of the IEEE international conference on computer vision workshops}, pages 554--561, 2013.

\bibitem[Krizhevsky et~al.(2009)Krizhevsky, Hinton, et~al.]{krizhevsky2009learning}
Alex Krizhevsky, Geoffrey Hinton, et~al.
\newblock Learning multiple layers of features from tiny images.
\newblock 2009.

\bibitem[Krizhevsky et~al.(2012)Krizhevsky, Sutskever, and Hinton]{krizhevsky2012imagenet}
Alex Krizhevsky, Ilya Sutskever, and Geoffrey~E Hinton.
\newblock Imagenet classification with deep convolutional neural networks.
\newblock \emph{Advances in neural information processing systems}, 25, 2012.

\bibitem[Kumar et~al.(2022)Kumar, Raghunathan, Jones, Ma, and Liang]{kumar2022finetuning}
Ananya Kumar, Aditi Raghunathan, Robbie~Matthew Jones, Tengyu Ma, and Percy Liang.
\newblock Fine-tuning can distort pretrained features and underperform out-of-distribution.
\newblock In \emph{International Conference on Learning Representations}, 2022.

\bibitem[Le and Yang(2015)]{le2015tiny}
Ya Le and Xuan Yang.
\newblock Tiny imagenet visual recognition challenge.
\newblock 2015.

\bibitem[Li et~al.(2022)Li, Li, Xiong, and Hoi]{li2022blip}
Junnan Li, Dongxu Li, Caiming Xiong, and Steven Hoi.
\newblock Blip: Bootstrapping language-image pre-training for unified vision-language understanding and generation.
\newblock In \emph{International conference on machine learning}, pages 12888--12900. PMLR, 2022.

\bibitem[Li et~al.(2023)Li, Li, Savarese, and Hoi]{li2023blip}
Junnan Li, Dongxu Li, Silvio Savarese, and Steven Hoi.
\newblock Blip-2: Bootstrapping language-image pre-training with frozen image encoders and large language models.
\newblock In \emph{International conference on machine learning}, pages 19730--19742. PMLR, 2023.

\bibitem[Li et~al.(2024)Li, Guan, Qiu, and Spratling]{li2024one}
Lin Li, Haoyan Guan, Jianing Qiu, and Michael Spratling.
\newblock One prompt word is enough to boost adversarial robustness for pre-trained vision-language models.
\newblock In \emph{Proceedings of the IEEE/CVF Conference on Computer Vision and Pattern Recognition}, pages 24408--24419, 2024.

\bibitem[Liao et~al.(2018)Liao, Liang, Dong, Pang, Hu, and Zhu]{liao2018defense}
Fangzhou Liao, Ming Liang, Yinpeng Dong, Tianyu Pang, Xiaolin Hu, and Jun Zhu.
\newblock Defense against adversarial attacks using high-level representation guided denoiser.
\newblock In \emph{Proceedings of the IEEE conference on computer vision and pattern recognition}, pages 1778--1787, 2018.

\bibitem[Liu et~al.(2023)Liu, Li, Wu, and Lee]{liu2023visual}
Haotian Liu, Chunyuan Li, Qingyang Wu, and Yong~Jae Lee.
\newblock Visual instruction tuning.
\newblock \emph{Advances in neural information processing systems}, 36:\penalty0 34892--34916, 2023.

\bibitem[L{\"u}ddecke and Ecker(2022)]{luddecke2022image}
Timo L{\"u}ddecke and Alexander Ecker.
\newblock Image segmentation using text and image prompts.
\newblock In \emph{Proceedings of the IEEE/CVF conference on computer vision and pattern recognition}, pages 7086--7096, 2022.

\bibitem[Madry et~al.(2018)Madry, Makelov, Schmidt, Tsipras, and Vladu]{madry2018towards}
Aleksander Madry, Aleksandar Makelov, Ludwig Schmidt, Dimitris Tsipras, and Adrian Vladu.
\newblock Towards deep learning models resistant to adversarial attacks.
\newblock In \emph{International Conference on Learning Representations}, 2018.

\bibitem[Maji et~al.(2013)Maji, Rahtu, Kannala, Blaschko, and Vedaldi]{maji2013fine}
Subhransu Maji, Esa Rahtu, Juho Kannala, Matthew Blaschko, and Andrea Vedaldi.
\newblock Fine-grained visual classification of aircraft.
\newblock \emph{arXiv preprint arXiv:1306.5151}, 2013.

\bibitem[Mao et~al.(2023)Mao, Geng, Yang, Wang, and Vondrick]{mao2023understanding}
Chengzhi Mao, Scott Geng, Junfeng Yang, Xin Wang, and Carl Vondrick.
\newblock Understanding zero-shot adversarial robustness for large-scale models.
\newblock In \emph{The Eleventh International Conference on Learning Representations}, 2023.

\bibitem[Mao et~al.(2024)Mao, Chen, Jia, Zhang, Xue, and Li]{mao2024context}
Xiaofeng Mao, Yufeng Chen, Xiaojun Jia, Rong Zhang, Hui Xue, and Zhao Li.
\newblock Context-aware robust fine-tuning.
\newblock \emph{International Journal of Computer Vision}, 132\penalty0 (5):\penalty0 1685--1700, 2024.

\bibitem[Nam et~al.(2024)Nam, Heo, and Lee]{nam2024lipsumft}
Giung Nam, Byeongho Heo, and Juho Lee.
\newblock Lipsum-{FT}: Robust fine-tuning of zero-shot models using random text guidance.
\newblock In \emph{The Twelfth International Conference on Learning Representations}, 2024.

\bibitem[Nilsback and Zisserman(2008)]{nilsback2008automated}
Maria-Elena Nilsback and Andrew Zisserman.
\newblock Automated flower classification over a large number of classes.
\newblock In \emph{2008 Sixth Indian conference on computer vision, graphics \& image processing}, pages 722--729. IEEE, 2008.

\bibitem[Oh et~al.(2024)Oh, Lim, Kim, Han, Yun, Choo, Hauptmann, Cheng, and Song]{oh2024towards}
Changdae Oh, Hyesu Lim, Mijoo Kim, Dongyoon Han, Sangdoo Yun, Jaegul Choo, Alexander Hauptmann, Zhi-Qi Cheng, and Kyungwoo Song.
\newblock Towards calibrated robust fine-tuning of vision-language models.
\newblock \emph{Advances in Neural Information Processing Systems}, 37:\penalty0 12677--12707, 2024.

\bibitem[Oord et~al.(2018)Oord, Li, and Vinyals]{oord2018representation}
Aaron van~den Oord, Yazhe Li, and Oriol Vinyals.
\newblock Representation learning with contrastive predictive coding.
\newblock \emph{arXiv preprint arXiv:1807.03748}, 2018.

\bibitem[Pan et~al.(2024)Pan, Li, and Yao]{Pan_Li_Yao_2024}
Chao Pan, Qing Li, and Xin Yao.
\newblock Adversarial initialization with universal adversarial perturbation: A new approach to fast adversarial training.
\newblock \emph{Proceedings of the AAAI Conference on Artificial Intelligence}, 38\penalty0 (19):\penalty0 21501--21509, 2024.

\bibitem[Parkhi et~al.(2012)Parkhi, Vedaldi, Zisserman, and Jawahar]{parkhi2012cats}
Omkar~M Parkhi, Andrea Vedaldi, Andrew Zisserman, and CV Jawahar.
\newblock Cats and dogs.
\newblock In \emph{2012 IEEE conference on computer vision and pattern recognition}, pages 3498--3505. IEEE, 2012.

\bibitem[Patashnik et~al.(2021)Patashnik, Wu, Shechtman, Cohen-Or, and Lischinski]{patashnik2021styleclip}
Or Patashnik, Zongze Wu, Eli Shechtman, Daniel Cohen-Or, and Dani Lischinski.
\newblock Styleclip: Text-driven manipulation of stylegan imagery.
\newblock In \emph{Proceedings of the IEEE/CVF international conference on computer vision}, pages 2085--2094, 2021.

\bibitem[Pratt et~al.(2023)Pratt, Covert, Liu, and Farhadi]{pratt2023does}
Sarah Pratt, Ian Covert, Rosanne Liu, and Ali Farhadi.
\newblock What does a platypus look like? generating customized prompts for zero-shot image classification.
\newblock In \emph{Proceedings of the IEEE/CVF international conference on computer vision}, pages 15691--15701, 2023.

\bibitem[Radford et~al.(2021)Radford, Kim, Hallacy, Ramesh, Goh, Agarwal, Sastry, Askell, Mishkin, Clark, et~al.]{radford2021learning}
Alec Radford, Jong~Wook Kim, Chris Hallacy, Aditya Ramesh, Gabriel Goh, Sandhini Agarwal, Girish Sastry, Amanda Askell, Pamela Mishkin, Jack Clark, et~al.
\newblock Learning transferable visual models from natural language supervision.
\newblock In \emph{International conference on machine learning}, pages 8748--8763. PMLR, 2021.

\bibitem[Rice et~al.(2020)Rice, Wong, and Kolter]{rice2020overfitting}
Leslie Rice, Eric Wong, and Zico Kolter.
\newblock Overfitting in adversarially robust deep learning.
\newblock In \emph{International conference on machine learning}, pages 8093--8104. PMLR, 2020.

\bibitem[Saha et~al.(2024)Saha, Van~Horn, and Maji]{saha2024improved}
Oindrila Saha, Grant Van~Horn, and Subhransu Maji.
\newblock Improved zero-shot classification by adapting vlms with text descriptions.
\newblock In \emph{Proceedings of the IEEE/CVF conference on computer vision and pattern recognition}, pages 17542--17552, 2024.

\bibitem[Sammani and Deligiannis(2024)]{sammani2024interpreting}
Fawaz Sammani and Nikos Deligiannis.
\newblock Interpreting and analysing clip's zero-shot image classification via mutual knowledge.
\newblock \emph{Advances in Neural Information Processing Systems}, 37:\penalty0 39597--39631, 2024.

\bibitem[Schlarmann et~al.(2024)Schlarmann, Singh, Croce, and Hein]{schlarmann2024robust}
Christian Schlarmann, Naman~Deep Singh, Francesco Croce, and Matthias Hein.
\newblock Robust clip: Unsupervised adversarial fine-tuning of vision embeddings for robust large vision-language models.
\newblock In \emph{International Conference on Machine Learning}, pages 43685--43704. PMLR, 2024.

\bibitem[Schuhmann et~al.(2021)Schuhmann, Kaczmarczyk, Komatsuzaki, Katta, Vencu, Beaumont, Jitsev, Coombes, and Mullis]{schuhmann2021laion}
Christoph Schuhmann, Robert Kaczmarczyk, Aran Komatsuzaki, Aarush Katta, Richard Vencu, Romain Beaumont, Jenia Jitsev, Theo Coombes, and Clayton Mullis.
\newblock Laion-400m: Open dataset of clip-filtered 400 million image-text pairs.
\newblock In \emph{NeurIPS Workshop Datacentric AI}, number FZJ-2022-00923. J{\"u}lich Supercomputing Center, 2021.

\bibitem[Sheng et~al.(2025)Sheng, Liang, Wang, and He]{sheng2025r}
Lijun Sheng, Jian Liang, Zilei Wang, and Ran He.
\newblock R-tpt: Improving adversarial robustness of vision-language models through test-time prompt tuning.
\newblock In \emph{Proceedings of the Computer Vision and Pattern Recognition Conference}, pages 29958--29967, 2025.

\bibitem[Szegedy et~al.(2014)Szegedy, Zaremba, Sutskever, Bruna, Erhan, Goodfellow, and Fergus]{szegedy2014intriguing}
Christian Szegedy, Wojciech Zaremba, Ilya Sutskever, Joan Bruna, Dumitru Erhan, Ian Goodfellow, and Rob Fergus.
\newblock Intriguing properties of neural networks.
\newblock In \emph{International Conference on Learning Representations}, 2014.

\bibitem[Tong et~al.(2025)Tong, Lai, Pan, and Yin]{tong2025zero}
Baoshun Tong, Hanjiang Lai, Yan Pan, and Jian Yin.
\newblock On the zero-shot adversarial robustness of vision-language models: A truly zero-shot and training-free approach.
\newblock In \emph{Proceedings of the Computer Vision and Pattern Recognition Conference}, pages 19921--19930, 2025.

\bibitem[Wang et~al.(2019)Wang, Ge, Lipton, and Xing]{wang2019learning}
Haohan Wang, Songwei Ge, Zachary Lipton, and Eric~P Xing.
\newblock Learning robust global representations by penalizing local predictive power.
\newblock \emph{Advances in neural information processing systems}, 32, 2019.

\bibitem[Wang et~al.(2025{\natexlab{a}})Wang, Chen, Li, Kang, Chen, and Tian]{wang2025declip}
Junjie Wang, Bin Chen, Yulin Li, Bin Kang, Yichi Chen, and Zhuotao Tian.
\newblock Declip: Decoupled learning for open-vocabulary dense perception.
\newblock In \emph{Proceedings of the Computer Vision and Pattern Recognition Conference}, pages 14824--14834, 2025{\natexlab{a}}.

\bibitem[Wang et~al.(2024)Wang, Zhang, Yuan, and Shan]{wang2024pre}
Sibo Wang, Jie Zhang, Zheng Yuan, and Shiguang Shan.
\newblock Pre-trained model guided fine-tuning for zero-shot adversarial robustness.
\newblock In \emph{Proceedings of the IEEE/CVF conference on computer vision and pattern recognition}, pages 24502--24511, 2024.

\bibitem[Wang et~al.(2025{\natexlab{b}})Wang, Chen, Zhang, Chen, and Ma]{wang2025tapt}
Xin Wang, Kai Chen, Jiaming Zhang, Jingjing Chen, and Xingjun Ma.
\newblock Tapt: Test-time adversarial prompt tuning for robust inference in vision-language models.
\newblock In \emph{Proceedings of the Computer Vision and Pattern Recognition Conference}, pages 19910--19920, 2025{\natexlab{b}}.

\bibitem[Waseda et~al.(2025)Waseda, Sugawara, and Echizen]{waseda2025quality}
Futa Waseda, Saku Sugawara, and Isao Echizen.
\newblock Quality text, robust vision: The role of language in enhancing visual robustness of vision-language models.
\newblock In \emph{Proceedings of the 33rd ACM International Conference on Multimedia}, pages 4808--4816, 2025.

\bibitem[Xiao et~al.(2010)Xiao, Hays, Ehinger, Oliva, and Torralba]{xiao2010sun}
Jianxiong Xiao, James Hays, Krista~A Ehinger, Aude Oliva, and Antonio Torralba.
\newblock Sun database: Large-scale scene recognition from abbey to zoo.
\newblock In \emph{2010 IEEE computer society conference on computer vision and pattern recognition}, pages 3485--3492. IEEE, 2010.

\bibitem[Xing et~al.(2025)Xing, Zhao, and Sebe]{xing2025clip}
Songlong Xing, Zhengyu Zhao, and Nicu Sebe.
\newblock Clip is strong enough to fight back: Test-time counterattacks towards zero-shot adversarial robustness of clip.
\newblock In \emph{Proceedings of the Computer Vision and Pattern Recognition Conference}, pages 15172--15182, 2025.

\bibitem[Yu et~al.(2024)Yu, Zhang, and Xu]{yu2024text}
Lu Yu, Haiyang Zhang, and Changsheng Xu.
\newblock Text-guided attention is all you need for zero-shot robustness in vision-language models.
\newblock \emph{Advances in Neural Information Processing Systems}, 37:\penalty0 96424--96448, 2024.

\bibitem[Zhang et~al.(2019)Zhang, Yu, Jiao, Xing, El~Ghaoui, and Jordan]{zhang2019theoretically}
Hongyang Zhang, Yaodong Yu, Jiantao Jiao, Eric Xing, Laurent El~Ghaoui, and Michael Jordan.
\newblock Theoretically principled trade-off between robustness and accuracy.
\newblock In \emph{International conference on machine learning}, pages 7472--7482. PMLR, 2019.

\bibitem[Zhang et~al.(2024)Zhang, Ma, Wang, Qiu, Wang, Jiang, and Sang]{zhang2024adversarial}
Jiaming Zhang, Xingjun Ma, Xin Wang, Lingyu Qiu, Jiaqi Wang, Yu-Gang Jiang, and Jitao Sang.
\newblock Adversarial prompt tuning for vision-language models.
\newblock In \emph{European conference on computer vision}, pages 56--72. Springer, 2024.

\bibitem[Zhang et~al.(2025)Zhang, Bi, Chen, Guo, and Cheng]{zhang2025clipure}
Mingkun Zhang, Keping Bi, Wei Chen, Jiafeng Guo, and Xueqi Cheng.
\newblock {CLIP}ure: Purification in latent space via {CLIP} for adversarially robust zero-shot classification.
\newblock In \emph{The Thirteenth International Conference on Learning Representations}, 2025.

\bibitem[Zhang et~al.(2022)Zhang, Guo, Zhang, Li, Miao, Cui, Qiao, Gao, and Li]{zhang2022pointclip}
Renrui Zhang, Ziyu Guo, Wei Zhang, Kunchang Li, Xupeng Miao, Bin Cui, Yu Qiao, Peng Gao, and Hongsheng Li.
\newblock Pointclip: Point cloud understanding by clip.
\newblock In \emph{Proceedings of the IEEE/CVF conference on computer vision and pattern recognition}, pages 8552--8562, 2022.

\bibitem[Zhao et~al.(2023)Zhao, Pang, Du, Yang, Li, Cheung, and Lin]{zhao2023evaluating}
Yunqing Zhao, Tianyu Pang, Chao Du, Xiao Yang, Chongxuan Li, Ngai-Man~Man Cheung, and Min Lin.
\newblock On evaluating adversarial robustness of large vision-language models.
\newblock \emph{Advances in Neural Information Processing Systems}, 36:\penalty0 54111--54138, 2023.

\bibitem[Zhong et~al.(2022)Zhong, Yang, Zhang, Li, Codella, Li, Zhou, Dai, Yuan, Li, et~al.]{zhong2022regionclip}
Yiwu Zhong, Jianwei Yang, Pengchuan Zhang, Chunyuan Li, Noel Codella, Liunian~Harold Li, Luowei Zhou, Xiyang Dai, Lu Yuan, Yin Li, et~al.
\newblock Regionclip: Region-based language-image pretraining.
\newblock In \emph{Proceedings of the IEEE/CVF conference on computer vision and pattern recognition}, pages 16793--16803, 2022.

\bibitem[Zhou et~al.(2022{\natexlab{a}})Zhou, Yang, Loy, and Liu]{zhou2022conditional}
Kaiyang Zhou, Jingkang Yang, Chen~Change Loy, and Ziwei Liu.
\newblock Conditional prompt learning for vision-language models.
\newblock In \emph{Proceedings of the IEEE/CVF conference on computer vision and pattern recognition}, pages 16816--16825, 2022{\natexlab{a}}.

\bibitem[Zhou et~al.(2022{\natexlab{b}})Zhou, Yang, Loy, and Liu]{zhou2022learning}
Kaiyang Zhou, Jingkang Yang, Chen~Change Loy, and Ziwei Liu.
\newblock Learning to prompt for vision-language models.
\newblock \emph{International Journal of Computer Vision}, 130\penalty0 (9):\penalty0 2337--2348, 2022{\natexlab{b}}.

\bibitem[Zhou et~al.(2024)Zhou, Xia, Lin, Han, and Liu]{zhou2024few}
Yiwei Zhou, Xiaobo Xia, Zhiwei Lin, Bo Han, and Tongliang Liu.
\newblock Few-shot adversarial prompt learning on vision-language models.
\newblock \emph{Advances in Neural Information Processing Systems}, 37:\penalty0 3122--3156, 2024.

\bibitem[Zhu et~al.(2024)Zhu, Chen, Shen, Li, and Elhoseiny]{zhuminigpt}
Deyao Zhu, Jun Chen, Xiaoqian Shen, Xiang Li, and Mohamed Elhoseiny.
\newblock Minigpt-4: Enhancing vision-language understanding with advanced large language models.
\newblock In \emph{The Twelfth International Conference on Learning Representations}, 2024.

\end{thebibliography}
}

\clearpage
\setcounter{page}{1}
\maketitlesupplementary


\section{More Dataset Information}
\label{App_sec:dataset_info}
In the main paper (\cref{tab:in_variants}), we employ several popular variants of ImageNet that share the pre-defined classes partially or entirely, but have distinctly different data domains, to reflect the limitations of a leveraging a large extensive dataset with labelled classes as a proxy. These variants include ImageNet-R \cite{hendrycks2021many}, ImageNet-A \cite{hendrycks2021natural}, ImageNet-S \cite{wang2019learning}, ObjectNet \cite{NEURIPS2019_97af07a1}:
\begin{itemize}
    \item \textbf{ImageNet-R}(endition) contains images of different renditions such as embroidery, paintings, toys, \etc.
    It has 30,000 images from 200 pre-defined classes, which is a subset of the 1,000 classes of ImageNet. 
    We use the textual prompt \texttt{`This is an artistic rendering of [CLS].'} when evaluating on this dataset.
    \item \textbf{ImageNet-A} contains natural image samples that standard models fail to classify. It has 7,500 images that belong to 200 pre-defined classes, which is a subset of the ImageNet classes. For evaluation, we employ the same textual prompt \texttt{'This is a photo of a [CLS].'} as in ImageNet.
    \item \textbf{ImageNet-Sketch} contains 50,000 images that are human-drawn black-and-white sketches. It has 1,000 pre-defined classes, which are the exact categories from ImageNet.
    For evaluation, we use the textual prompt of \texttt{`This is a sketch of a [CLS].'}.
    \item \textbf{ObjectNet} includes 50k real photographs of objects unusually arranged, such as varied camera angles, object poses, and diverse backgrounds. It has a total of 313 classes, of which 113 classes overlap with ImageNet.
    For evaluation, we employ the same textual prompt \texttt{'This is a photo of a [CLS].'} as in ImageNet.
\end{itemize}

\section{Other Ablation Studies}
\label{App_sec:more_ablations}
In the main paper, we conduct ablative studies on the regularisation terms (\cref{sec:ablation_regularisation}).
In this section, we perform ablative studies on other training settings.

\subsection{Data Amount and Batch Size}
\begin{figure}
    \centering
    \includegraphics[width=1.\linewidth]{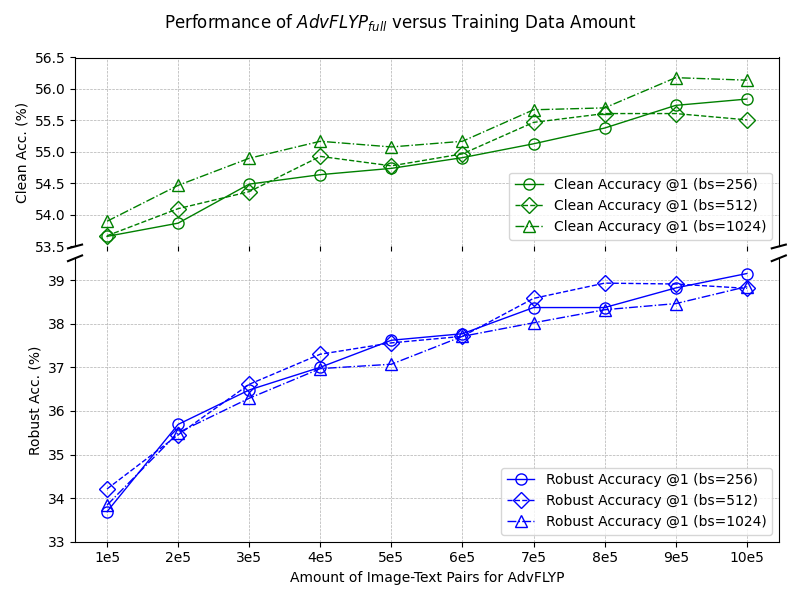}
    \caption{Performance of $\textrm{AdvFLYP}_{full}$ averaged over 14 downstream datasets versus the amount of image-text pairs from the web. The robust accuracy is evaluated under PGD-10 ($\epsilon=1/255$).}
    \label{fig:training_data_amount}
\end{figure}
We implement $\textrm{AdvFLYP}_{full}$ with varying amounts of image-text pairs and three batch sizes (256, 512 and 1024), and evaluate the performance of the finetuned model on 14 downstream datasets. 
\cref{fig:training_data_amount} reports the results.
It can be seen that both adversarial robustness and clean accuracy steadily increase as the model is finetuned on an increasingly large amount of image-text pairs.
In the main paper, we employ one million noisy image-text pairs collected from the web and fix the training data amount to 1M to reduce training time. 
The performance of $\textrm{AdvFLYP}_{full}$ is projected to continue to further improve if we enlarge the training data amount further.
As opposed to previous AFT methods that finetune CLIP via a cross-entropy loss, where the batch size does not play an important role, 
AdvFLYP is a contrastive finetuning paradigm, where samples are contrasted with each other in a batch, and a large batch size benefits model performance because it provides more negative examples \cite{radford2021learning}.
The original paper of CLIP adopts the batch size of around 32k during pretraining to guarantee sufficient negative examples within a single batch.
In this work, we are not able to adopt the same batch size for adversarial finetuning due to hardware constraints.
Nonetheless, we observe a different pattern of impact of the batch size in our adversarial contrastive finetuning paradigm.
Experimentally, we find that a batch size smaller than 256 leads to poor performance in terms of robustness and clean accuracy on downstream datasets, which implies that a sufficiently large batch size is also crucial for effective contrastive learning in our AdvFLYP paradigm.
However, when the batch size is further increased (512 and 1024), there is a trade-off between robustness and clean accuracy, with a larger batch size benefiting the clean accuracy while suppressing the robustness gains, to some extent.
One possible reason is that for a larger batch with more examples, it is more demanding to optimise the batch-wise perturbations (\cref{eq:flyp_attack}) that maximise the contrastive loss, leading to less difficult adversarial examples.
Therefore, finetuning $f_\theta$ based on these examples leads to lesser robustness gains, while better retaining the zero-shot capabilities on clean images.
In the main paper, we fix the batch size at 256.

\begin{table}[t]
    \centering
    \begin{tabular}{c|c|c|c||c}
    \toprule
    (\%) & PGD & CW & AA & \textit{clean} \\ 
    \midrule
    $f_\theta$, $g_\phi$ &35.67 &34.72 &33.89  &53.81 \\
    \textit{all} &35.71 &34.63 &33.76 &53.97 \\
    \midrule
    \rowcolor{blue!5}
    $f_\theta$ (ours) &\bf39.15 &\bf38.30 &\bf37.74 &\bf55.84 \\
    \bottomrule
    \end{tabular}
    \caption{We unfreeze more modules of CLIP in our $\textrm{AdvFLYP}_{full}$. We evaluate the finetuned models using the attack methods of PGD, CW, and AutoAttack with $\epsilon=1/255$.}
    \label{tab:more_trainable_modules}
\end{table}

\subsection{More Tunable Modules}
Despite the fact that we stick to the training recipe (mainly the training data distribution and training objective) of CLIP's pretraining, 
AdvFLYP differs from the pretraining process in that it only finetunes the pre-trained vision encoder $f_\theta$ of CLIP, and keeps all other modules of CLIP frozen.
In this section, we unfreeze more modules of CLIP and implement AdvFLYP.
\cref{tab:more_trainable_modules} reports the  performance of $\textrm{AdvFLYP}_{full}$ when we finetune the text encoder $g_\phi$, and all modules in addition to the vision encoder $f_\theta$.
It can be seen that both variants degrade the performance of $\textrm{AdvFLYP}_{full}$ significantly, which indicates that the vision encoder $f_\theta$ is the component of central importance to robustfying CLIP.
We speculate that the degradation is due to the unnecessary distortion of the text encoder $g_\phi$.
Considering that the adversarial images are only fed to the vision encoder $f_\theta$, intuitively, one should only finetune $f_\theta$ to ensure that these adversarial images are aligned with the correct text supervision signals.
Therefore, finetuning $g_\phi$ and other modules in addition to the vision encoder $f_\theta$ does not benefit the overall performance.

\subsection{Regularisation Formulation}
\label{sec:regularisation_2nd_terms}
\begin{table}[t]
    \centering
    \begin{tabular}{c|c|c||c}
    \toprule
    (\%) & PGD & AA & clean \\
    \midrule
    $\textrm{AdvFLYP}_{2nd}$ & 32.71 & 31.25 & 49.01\\ 
    \rowcolor{blue!5}$\textrm{AdvFLYP}_{full}$ & \textbf{33.69} & \textbf{31.79} & \textbf{53.66} \\
    \bottomrule
    \end{tabular}
    \caption{Adversarial robustness evaluated at $\epsilon=1/255$ and clean accuracy averaged over 14 datasets. Both variants are trained with 100k image-text pairs.}
    \label{tab:reg_2nd_terms}
\end{table}
In the main paper, we formulate regularisation by computing (i) the deviation between adversarial and clean image features by the target model $f_\theta$, and (ii) the deviation between the adversarial image features by the target model $f_\theta$ and the original CLIP $F_{\theta_0}$.
This applies to both logit-level (\cref{eq:logit_reg}) and feature-level (\cref{eq:feat_reg}) regularisation.
Prior work on adversarial defence has employed the second term (ii) to defend neural networks \cite{liao2018defense}.
As a preliminary experiment, we evaluate whether using the second term (ii) of \cref{eq:logit_reg} and \cref{eq:feat_reg} suffices to boost adversarial robustness of the target model. 
Specifically, we randomly sample 100k training data, on which we perform regularised AdvFLYP with only the (ii) terms for both logit and feature levels, and denote this variant as $\textrm{AdvFLYP}_{2nd}$. The results for this variant are reported in \cref{tab:reg_2nd_terms}.
From the table, it can be concluded that the first term (i) is also important, especially for the retention of clean accuracy, hence the complete formulation of the proposed regularisation in the main paper.

\section{More Experimental Results}
\label{App_sec:more_experiment_results}

\begin{table}[t]
    \centering
    \begin{tabular}{c|c|c|c||c}
    \toprule
    (\%) &PGD &CW  &AA  &\textit{clean} \\
    \midrule
    TeCoA &\bf33.77 &\bf33.04 &\bf32.17 &51.35 \\
    naive FLYP &32.89 &32.28 &31.23 &\bf51.66 \\
    \midrule
    PMG-AFT&\bf35.85 &\bf34.87 &\bf33.89 &\bf54.25 \\
    naive FLYP + $\mathcal{L}_{logit}$&35.28  &34.35  &33.33  &53.87 \\
    \bottomrule
    \end{tabular}
    \caption{We implement `naive FLYP' in our adversarial finetuning context and compare to the previous standard AFT paradigm TeCoA \cite{mao2023understanding}. PMG-AFT \cite{wang2024pre} is equivalent to TeCoA + $\mathcal{L}_{logit}$.
    We evaluate the robustness of the finetuned models using PGD, CW, and AutoAttack with $\epsilon=1/255$ and report the average results over 14 downstream datasets.
    }
    \label{tab:naive_FLYP}
\end{table}

\begin{table}[t]
    \centering
    \resizebox{\linewidth}{!}{
    \begin{tabular}{c|c|cc|c}
    \toprule
    \multicolumn{2}{c|}{(\%)} & \begin{tabular}{@{}c@{}}AutoAttack \\ ($\epsilon=1/255$)\end{tabular} & Clean & Avg.\\
    \midrule
    & \textcolor{lightgray}{CLIP} & \textcolor{lightgray}{0.05} & \textcolor{lightgray}{62.03} & 
    \textcolor{lightgray}{31.04} \\
    \cmidrule{2-5}
    (a) &TeCoA & 32.17 & 51.35 & 41.76 \\
    (b) &TeCoA + $\mathcal{L}_{logit}$& \bf33.89 & \bf54.25 & \bf44.07\\
    (c) &TeCoA + $\mathcal{L}_{feat}$&32.94 &52.31  &42.63 \\
    (d) &TeCoA + $\mathcal{L}_{logit}$ + $\mathcal{L}_{feat}$& 33.88 &54.22 &44.05 \\
    \bottomrule
    \end{tabular}
    }
    \caption{Results of the models finetuned with different combinations of regularisation levels on top of TeCoA \cite{mao2023understanding}. The combination (\textit{b}) is equivalent to PMG-AFT \cite{wang2024pre}. The reported results are the average accuracy over 14 downstream datasets.
    }
    \label{tab:tecoa_reg_ablation}
\end{table}

\begin{table*}[t]
    \centering
    \resizebox{\textwidth}{!}{
    \begin{tabular}{c|c|cccccccccccccc|c}
    \toprule
    \multicolumn{2}{c|}{\raisebox{3\height}{(\%)}} 
    & \spin{CIFAR10} & \spin{CIFAR100} & \spin{STL10} & \spin{Caltech101} & \spin{Caltech256} & \spin{OxfordPets} & \spin{Flowers102} & \spin{Food101} & \spin{StanfordCars} & \spin{SUN397} & \spin{Country211} & \spin{FGVCAircraft} & \spin{EuroSAT} & \spin{DTD} & \spin{avg.}\\
    \midrule\midrule
    \multirow{7}{*}{\spin{PGD}}
    & \color{lightgray}CLIP 
    & \color{lightgray}0.00 & \color{lightgray}0.00 &\color{lightgray}0.71 &\color{lightgray}1.71 &\color{lightgray}0.84 &\color{lightgray}0.00 &\color{lightgray}0.02 &\color{lightgray}0.01 &\color{lightgray}0.00 &\color{lightgray}0.02 &\color{lightgray}0.00 &\color{lightgray}0.00 &\color{lightgray}0.00 &\color{lightgray}0.21 &\color{lightgray}0.25 \\
    & FARE 
    &1.11 &0.85 &15.21 &22.14 &14.02 &3.41 &1.12 &0.72 &0.15 &1.79 &0.04 &0.00 &2.69 &5.11 &4.88  \\
    & TeCoA 
    &11.66 &6.75 &46.08 &51.03 &37.80 &\un{31.32} &14.15 &9.03 &3.44 &13.14 &0.92 &1.08 &\un{9.85} &14.20 &17.89  \\
    & PMG-AFT 
    &12.75 &7.68 &47.84 &51.15 &38.41 &\bf32.98 &15.09 &10.34 &4.42 &13.29 &0.89 &1.59 &\bf10.31 &14.52 &18.66  \\
    & TGA-ZSR 
    &2.99 &1.34 &27.29 &33.30 &23.64 &19.13 &4.15 &4.26 &1.08 &5.67 &0.21 &0.09 &0.56 &6.60 &9.31  \\
    \cmidrule{2-17}
    & 
    AdvFLYP 
    &\un{19.32} &\un{9.63} &\un{52.24} &\un{52.96} &\un{41.13} &27.61 &\un{18.13} &\un{12.40} &\un{9.15} &\un{16.51} &\un{0.99} &\un{1.80} &0.60 &\bf16.70 &\un{19.94}  \\
    & 
    $\mathrm{AdvFLYP}_{full}$ 
    &\bf26.15 &\bf13.27 &\bf57.28 &\bf55.23 &\bf42.99 &29.54 &\bf18.95 &\bf14.54 &\bf9.35 &\bf17.48 &\bf1.08 &\bf1.92 &3.88 &\un{16.65} &\bf22.02  \\
    \midrule\midrule
    
    \multirow{7}{*}{\spin{CW}}
    & \color{lightgray}CLIP 
    & \color{lightgray}0.00 & \color{lightgray}0.00 &\color{lightgray}0.72 &\color{lightgray}6.33 &\color{lightgray}0.71 &\color{lightgray}0.00 &\color{lightgray}0.00 &\color{lightgray}0.01 &\color{lightgray}2.30 &\color{lightgray}0.02 &\color{lightgray}0.00 &\color{lightgray}0.00 &\color{lightgray}0.00 &\color{lightgray}0.37 &\color{lightgray}0.75 \\
    & FARE 
    &1.36 &1.16 &17.02 &28.48 &16.15 &3.82 &1.50 &1.15 &2.52 &2.56 &0.06 &0.00 &2.34 &4.63 &5.91  \\
    & TeCoA 
    &12.20 &6.96 &47.42 &54.35 &39.23 &\un{33.66} &13.95 &9.82 &4.84 &14.40 &0.96 &1.41 &\un{9.23} &12.98 &18.67  \\
    & PMG-AFT 
    &12.46 &7.39 &49.05 &55.77 &39.89 &\bf34.97 &13.50 &11.56 &5.55 &14.11 &0.87 &1.14 &\bf10.13 &13.14 &19.25  \\
    & TGA-ZSR 
    &3.37 &1.68 &28.49 &37.71 &24.45 &18.64 &4.94 &4.96 &3.11 &6.69 &0.34 &0.27 &0.54 &5.90 &10.08  \\
    \cmidrule{2-17}
    & 
    AdvFLYP 
    &\un{20.07} &\un{10.44} &\un{53.31} &\un{57.20} &\un{42.68} &30.53 &\bf18.78 &\un{14.51} &\bf12.19 &\un{17.99} &\un{1.12} &\bf2.13 &0.63 &\bf16.22 &\un{21.27}  \\
    & 
    $\mathrm{AdvFLYP}_{full}$ 
    &\bf24.64 &\bf12.30 &\bf57.89 &\bf58.19 &\bf44.12 &32.43 &\un{18.56} &\bf16.81 &\un{11.33} &\bf18.34 &\bf1.13 &\un{1.92} &4.69 &\un{15.37} &\bf22.69  \\
    \midrule\midrule
    
    \multirow{7}{*}{\spin{AutoAttack}}
    & \color{lightgray}CLIP 
    & \color{lightgray}0.00 & \color{lightgray}0.06 &\color{lightgray}0.00 &\color{lightgray}0.05 &\color{lightgray}0.01 &\color{lightgray}0.00 &\color{lightgray}0.02 &\color{lightgray}0.00 &\color{lightgray}0.00 &\color{lightgray}0.00 &\color{lightgray}0.00 &\color{lightgray}0.03 &\color{lightgray}0.08 &\color{lightgray}0.05 &\color{lightgray}0.02 \\
    & FARE 
    &0.35 &0.57 &9.72 &17.07 &10.86 &1.72 &0.75 &0.49 &0.14 &0.97 &0.01 &0.03 &1.97 &3.30 &3.42  \\
    & TeCoA 
    &9.97 &5.83 &44.46 &49.31 &36.09 &\un{29.63} &12.44 &7.58 &2.71 &11.26 &0.66 &0.72 &\un{8.97} &12.39 &16.57  \\
    & PMG-AFT 
    &10.34 &6.17 &45.92 &49.64 &36.53 &\bf30.64 &12.15 &8.60 &3.00 &10.98 &0.59 &0.87 &\bf9.86 &12.77 &17.00  \\
    & TGA-ZSR 
    &0.00 &0.00 &0.01 &0.06 &0.02 &0.00 &0.03 &0.00 &0.00 &0.01 &0.01 &0.00 &0.10 &0.05 &0.02  \\
    \cmidrule{2-17}
    & 
    AdvFLYP 
    &\un{17.87} &\un{8.88} &\un{51.29} &\un{51.91} &\un{39.96} &26.38 &\bf17.01 &\un{11.33} &\bf8.13 &\un{15.10} &\bf0.78 &\bf1.32 &0.34 &\bf15.85 &\un{19.01}  \\
    & 
    $\mathrm{AdvFLYP}_{full}$ 
    &\bf23.02 &\bf11.18 &\bf56.22 &\bf53.90 &\bf41.29 &27.42 &\un{16.95} &\bf13.18 &\un{7.72} &\bf15.33 &\un{0.74} &\un{1.29} &1.83 &\un{14.84} &\bf20.35  \\
    \midrule\midrule
    
    \multirow{7}{*}{\spin{$\overline{\mathrm{AVG}}$}}
    & \color{lightgray}CLIP 
    & \color{lightgray}0.00 & \color{lightgray}0.02 &\color{lightgray}0.48 &\color{lightgray}2.69 &\color{lightgray}0.52 &\color{lightgray}0.00 &\color{lightgray}0.01 &\color{lightgray}0.01 &\color{lightgray}0.77 &\color{lightgray}0.01 &\color{lightgray}0.00 &\color{lightgray}0.01 &\color{lightgray}0.03 &\color{lightgray}0.21 &\color{lightgray}0.34 \\
    & FARE 
    &0.94 &0.86 &13.99 &22.56 &13.68 &2.98 &1.12 &0.79 &0.94 &1.77 &0.04 &0.01 &2.33 &4.34 &4.74  \\
    & TeCoA 
    &11.28 &6.51 &45.99 &51.57 &37.71 &\un{31.53} &13.51 &8.81 &3.66 &12.94 &0.85 &1.07 &\un{9.35} &13.19 &17.71  \\
    & PMG-AFT 
    &11.85 &7.08 &47.60 &52.18 &38.27 &\bf32.86 &13.58 &10.17 &4.32 &12.80 &0.78 &1.20 &\bf10.10 &13.48 &18.31  \\
    & TGA-ZSR 
    &2.12 &1.01 &18.60 &23.69 &16.04 &12.59 &3.04 &3.07 &1.40 &4.12 &0.19 &0.12 &0.40 &4.18 &6.47  \\
    \cmidrule{2-17}
    & 
    AdvFLYP 
    &\un{19.09} &\un{9.65} &\un{52.28} &\un{54.02} &\un{41.26} &28.17 &\un{17.98} &\un{12.74} &\bf9.82 &\un{16.53} &\un{0.96} &\bf1.75 &0.52 &\bf16.26 &\un{20.07}  \\
    &
    $\mathrm{AdvFLYP}_{full}$ 
    &\bf24.60 &\bf12.25 &\bf57.13 &\bf55.77 &\bf42.80 &29.80 &\bf18.15 &\bf14.84 &\un{9.47} &\bf17.05 &\bf0.98 &\un{1.71} &3.47 &\un{15.62} &\bf21.69  \\
    \bottomrule 
    \end{tabular}
    }
    \vspace{-.7em}
    \caption{Classification accuracy (\%) 
    on 14 downstream datasets tested with three adversarial attack algorithms ($\epsilon=2/255$).
    We highlight the \textbf{best} and \underline{second best} result.
    }
    \label{tab:test_eps2}
\end{table*}

\begin{table*}[t] 
    \centering
    \resizebox{\textwidth}{!}{
    \begin{tabular}{c|c|cccccccccccccc|c}
    \toprule
    \multicolumn{2}{c|}{\raisebox{3\height}{(\%)}} 
    & \spin{CIFAR10} & \spin{CIFAR100} & \spin{STL10} & \spin{Caltech101} & \spin{Caltech256} & \spin{OxfordPets} & \spin{Flowers102} & \spin{Food101} & \spin{StanfordCars} & \spin{SUN397} & \spin{Country211} & \spin{FGVCAircraft} & \spin{EuroSAT} & \spin{DTD} & \spin{avg.}\\
    \midrule\midrule
    \multirow{7}{*}{\spin{PGD}}
    & \color{lightgray}CLIP 
    & \color{lightgray}0.00 & \color{lightgray}0.00 &\color{lightgray}0.04 &\color{lightgray}0.61 &\color{lightgray}0.11 &\color{lightgray}0.00 &\color{lightgray}0.00 &\color{lightgray}0.00 &\color{lightgray}0.00 &\color{lightgray}0.00 &\color{lightgray}0.00 &\color{lightgray}0.00 &\color{lightgray}0.00 &\color{lightgray}0.11 &\color{lightgray}0.06 \\
    & FARE 
    &0.06 &0.02 &1.42 &5.05 &2.04 &1.55 &0.02 &0.02 &0.00 &0.03 &0.00 &0.00 &0.00 &0.37 &0.76  \\
    & TeCoA 
    &0.72 &0.79 &9.39 &20.42 &12.46 &\bf3.08 &2.00 &0.63 &0.06 &2.06 &\bf0.10 &0.00 &\un{5.25} &4.63 &4.40  \\
    & PMG-AFT 
    &0.53 &0.89 &9.91 &20.02 &12.24 &2.81 &1.74 &0.76 &0.06 &1.92 &0.07 &\bf0.03 &\bf6.34 &4.84 &4.44  \\
    & TGA-ZSR 
    &0.04 &0.01 &2.10 &6.37 &3.47 &2.10 &0.05 &0.10 &0.00 &0.22 &0.01 &0.00 &0.00 &0.80 &1.09  \\
    \cmidrule{2-17}
    & 
    AdvFLYP 
    &\un{2.08} &\un{1.92} &\un{19.50} &\bf27.69 &\un{17.20} &\un{2.94} &\bf3.27 &\un{1.23} &\bf0.85 &\bf3.30 &0.06 &\bf0.03 &0.01 &\un{6.44} &\un{6.18}  \\
    & 
    $\mathrm{AdvFLYP}_{full}$ 
    &\bf2.91 &\bf2.97 &\bf22.19 &\un{27.28} &\bf17.29 &2.73 &\un{2.91} &\bf1.34 &\un{0.57} &\un{3.11} &\un{0.08} &0.00 &0.07 &\bf6.81 &\bf6.45\\
    \midrule\midrule
    
    \multirow{7}{*}{\spin{CW}}
    & \color{lightgray}CLIP 
    & \color{lightgray}0.00 & \color{lightgray}0.00 &\color{lightgray}0.04 &\color{lightgray}4.94 &\color{lightgray}0.07 &\color{lightgray}0.00 &\color{lightgray}0.00 &\color{lightgray}0.00 &\color{lightgray}2.41 &\color{lightgray}0.00 &\color{lightgray}0.00 &\color{lightgray}0.00 &\color{lightgray}0.00 &\color{lightgray}0.11 &\color{lightgray}0.54 \\
    & FARE 
    &0.01 &0.08 &1.55 &8.15 &1.87 &0.03 &0.00 &0.02 &2.19 &0.04 &0.00 &0.00 &0.00 &0.32 &1.02  \\
    & TeCoA 
    &0.80 &1.00 &11.02 &25.58 &14.50 &\bf4.80 &2.19 &0.87 &1.52 &3.11 &\bf0.14 &0.03 &\un{5.54} &4.10 &5.37  \\
    & PMG-AFT 
    &0.43 &0.98 &10.66 &25.85 &14.14 &4.58 &1.61 &0.93 &1.74 &2.71 &\un{0.13} &0.03 &\bf6.75 &3.72 &5.30  \\
    & TGA-ZSR 
    &0.05 &0.05 &1.86 &8.61 &3.12 &0.33 &0.05 &0.08 &1.60 &0.33 &0.00 &0.00 &0.00 &0.48 &1.18  \\
    \cmidrule{2-17}
    & 
    AdvFLYP 
    &\un{2.19} &\un{2.32} &\un{20.36} &\bf33.55 &\bf18.97 &4.42 &\bf3.80 &\un{1.76} &\bf3.50 &\bf4.36 &0.11 &\bf0.12 &0.00 &\bf6.33 &\un{7.27}  \\
    & 
    $\mathrm{AdvFLYP}_{full}$ 
    &\bf2.51 &\bf2.94 &\bf22.89 &\un{33.54} &\un{18.65} &\un{4.74} &\un{3.40} &\bf1.97 &\un{2.72} &\un{4.09} &0.12 &\un{0.09} &0.05 &\un{5.32} &\bf7.36  \\
    \midrule\midrule
    
    \multirow{7}{*}{\spin{AutoAttack}}
    & \color{lightgray}CLIP 
    & \color{lightgray}0.00 & \color{lightgray}0.06 &\color{lightgray}0.00 &\color{lightgray}0.00 &\color{lightgray}0.00 &\color{lightgray}0.00 &\color{lightgray}0.02 &\color{lightgray}0.00 &\color{lightgray}0.00 &\color{lightgray}0.00 &\color{lightgray}0.00 &\color{lightgray}0.03 &\color{lightgray}0.08 &\color{lightgray}0.05 &\color{lightgray}0.02 \\
    & FARE 
    &0.00 &0.04 &0.00 &1.30 &0.37 &0.00 &0.02 &0.00 &0.00 &0.00 &0.00 &\bf0.03 &0.04 &0.11 &0.14  \\
    & TeCoA 
    &0.10 &0.24 &4.28 &14.34 &8.59 &\bf1.64 &1.06 &0.28 &0.11 &0.92 &\bf0.05 &0.00 &0.08 &3.14 &2.49  \\
    & PMG-AFT 
    &0.04 &0.26 &3.76 &13.39 &7.56 &1.14 &0.80 &0.31 &0.01 &0.76 &\bf0.05 &\bf0.03 &\bf0.38 &3.14 &2.26  \\
    & TGA-ZSR 
    &0.00 &0.00 &0.00 &0.02 &0.00 &0.00 &0.03 &0.00 &0.00 &0.01 &0.01 &0.00 &\un{0.10} &0.05 &0.02  \\
    \cmidrule{2-17}
    & 
    AdvFLYP 
    &\un{0.50} &\un{0.91} &\un{10.98} &\bf20.96 &\bf12.78 &\un{1.55} &\bf2.28 &\bf0.65 &\bf0.51 &\bf1.67 &\bf0.05 &\bf0.03 &0.01 &\bf5.27 &\bf4.15  \\
    & 
    $\mathrm{AdvFLYP}_{full}$ 
    &\bf0.65 &\bf1.42 &\bf12.65 &\un{20.02} &\un{11.46} &1.09 &\un{1.76} &\un{0.63} &\un{0.26} &\un{1.30} &0.04 &0.00 &0.04 &\un{4.47} &\un{3.98}  \\
    \midrule\midrule
    
    \multirow{7}{*}{\spin{$\overline{\mathrm{AVG}}$}}
    & \color{lightgray}CLIP 
    & \color{lightgray}0.00 & \color{lightgray}0.02 &\color{lightgray}0.02 &\color{lightgray}1.85 &\color{lightgray}0.06 &\color{lightgray}0.00 &\color{lightgray}0.01 &\color{lightgray}0.00 &\color{lightgray}0.80 &\color{lightgray}0.00 &\color{lightgray}0.00 &\color{lightgray}0.01 &\color{lightgray}0.03 &\color{lightgray}0.09 &\color{lightgray}0.21 \\
    & FARE 
    &0.02 &0.05 &0.99 &4.83 &1.42 &0.53 &0.01 &0.01 &0.73 &0.03 &0.00 &0.01 &0.01 &0.27 &0.64  \\
    & TeCoA 
    &0.54 &0.68 &8.23 &20.11 &11.85 &\bf3.17 &1.75 &0.59 &0.56 &2.03 &\bf0.10 &0.01 &\un{3.62} &3.95 &4.09  \\
    & PMG-AFT 
    &0.33 &0.71 &8.11 &19.75 &11.31 &2.84 &1.38 &0.67 &0.61 &1.80 &\un{0.08} &\un{0.03} &\bf4.49 &3.90 &4.00  \\
    & TGA-ZSR 
    &0.03 &0.02 &1.32 &5.00 &2.20 &0.81 &0.04 &0.06 &0.53 &0.19 &0.01 &0.00 &0.03 &0.44 &0.76  \\
    \cmidrule{2-17}
    & 
    AdvFLYP 
    &\un{1.59} &\un{1.72} &\un{16.95} &\bf27.40 &\bf16.32 &\un{2.97} &\bf3.12 &\un{1.21} &\bf1.62 &\bf3.11 &0.07 &\bf0.06 &0.01 &\bf6.01 &\un{5.87}  \\
    & 
    $\mathrm{AdvFLYP}_{full}$ 
    &\bf2.02 &\bf2.44 &\bf19.24 &\un{26.94} &\un{15.80} &2.85 &\un{2.69} &\bf1.31 &\un{1.19} &\un{2.84} &\un{0.08} &\un{0.03} &0.06 &\un{5.53} &\bf5.93  \\
    \bottomrule 
    \end{tabular}
    }
    \vspace{-.7em}
    \caption{Classification accuracy (\%) 
    on 14 downstream datasets tested with three adversarial attack algorithms ($\epsilon=4/255$).
    We highlight the \textbf{best} and \underline{second best} result.
    }
    \label{tab:test_eps4}
\end{table*}

\subsection{Robustness under Higher Attack Budgets}
We report the full tables of robustness evaluated under the attack strength of $\epsilon=2/255$ and $\epsilon=4/255$ in \cref{tab:test_eps2} and \cref{tab:test_eps4}, respectively.
It can be seen that our AdvFLYP still consistently outperforms previous the previous AFT paradigm TeCoA \cite{mao2023understanding} and its regularisation-based advancements PMG-AFT \cite{wang2024pre} and TGA-ZSR \cite{yu2024text} under strong attack budgets, showing the reliability of our simple paradigm.
In contrast, the method that formulates regularisation based on text-guided attention (TGA-ZSR \cite{yu2024text}) is shown to be less effective on higher attack strengths, implying that it may have overfit to the attack strength ($\epsilon=1/255$) during adversarial finetuning.
On average, under the attack budget of $\epsilon=2/255$, our basic AdvFLYP paradigm and its regularised variant $\textrm{AdvFLYP}_{full}$ achieve the robust accuracy of $20.07\%$ and $21.69\%$, outperforming PMG-AFT ($18.31\%$) by a relative margin of $9.61\%$ and $18.46\%$ over 14 downstream datasets and various attack methods, respectively.
When evaluated under the budget of $\epsilon=4/255$, AdvFLYP and $\textrm{AdvFLYP}_{full}$ achieve an average robustness of $5.87\%$ and $5.93\%$, 
both outperforming PMG-AFT ($4.00\%$).
Results show that the paradigm of AdvFLYP is a competitive adversarial finetuning paradigm for VLMs despite its sheer simplicity, compared to the standard practice of finetuning VLMs on a large and extensive dataset with labelled classes such as ImageNet.

\subsection{Naive FLYP for AFT}
\label{sec:naive_flyp}
This work draws inspiration from FLYP \cite{goyal2023finetune}, which finds that finetuning CLIP with a contrastive loss as employed in the pretraining process helps to improve generalisation to \textit{out-of-distribution} (OOD) data.
FLYP challenges previous standard finetuning practices that finetune CLIP with a cross-entropy loss on \textit{in-distribution} (ID) data, and leverages a contrastive loss instead.
However, they still operate on classification-oriented ID data, and ignore the overlap of classes present in a batch.
In contrast, AdvFLYP aims to underscore the importance of following the data distribution and the training objective of VLMs' pretraining to boost its zero-shot adversarial robustness.
Therefore, despite the fact that AdvFLYP stems from the same spirit as FLYP, its motivation and implementation differ fundamentally from those of FLYP, and is no simple extension of FLYP in the context of adversarial robustness.
In this section, we naively apply FLYP for adversarial finetuning.
Specifically, we finetune CLIP's vision encoder $f_\theta$ on ImageNet. Instead of employing a cross-entropy loss as in TeCoA \cite{mao2023understanding} and PMG-AFT \cite{wang2024pre}, we follow the implementation of FLYP and leverage the contrastive loss. 
As in FLYP, we ignore the fact that some classes may overlap in one batch.
We report the results in \cref{tab:naive_FLYP}.
It can be seen that performing FLYP naively does not lead to improved robustness or better clean accuracy in the context of adversarial finetuning.
In contrast, by performing real contrastive finetuning on adversarial web images and their corresponding texts, our AdvFLYP paradigm achieves significantly improved robustness compared to previous AFT methods.

\subsection{Other Vision Backbones}\label{sec:clip-vit-b16}
\begin{table}[t]
    \centering
    \begin{tabular}{c|c|c||c}
    \toprule
    (\%) & PGD & AA & clean \\
    \midrule
    PMG-AFT &31.59 & 28.92 &54.46 \\
    \rowcolor{blue!5}$
    \textrm{AdvFLYP}_{full}$ &\textbf{35.13} & \textbf{33.62} &\textbf{56.07} \\ 
    \bottomrule
    \end{tabular}
    \caption{Robustness ($\epsilon=1/255$) and clean accuracy of CLIP ViT-B/16 averaged over 14 datasets.}
    \label{tab:clip_vit16}
\end{table}
In the main paper, we focus on the CLIP ViT-B/32 model, following the practice of prior work \cite{mao2023understanding,wang2024pre,yu2024text}.
The AdvFLYP paradigm can be readily employed to boost the adversarial robustness of other CLIP backbones and other CLIP-style VLMs.
In this section, we conduct a preliminary experiment on CLIP ViT-B/16 with 100k image-text pairs to show that AdvFLYP achieves consistent improvement over previous paradigms. Results reported in \cref{tab:clip_vit16} show that the AdvFLYP paradigm consistently outperforms previous AFT paradigms on other vision backbones.

\section{Training Data Analysis}
\label{App_sec:training_data}
The current \textit{de facto} standard paradigm for finetuning VLMs to achieve zero-shot adversarial robustness is largely based on the adversarial training (AT) principles of classical adversarial learning \cite{madry2018towards}, which involve a dataset of labelled classes. 
This paradigm is reasonable in the sense that the finetuned CLIP is to be deployed in downstream classification datasets.
However, we believe that adversarial finetuning should not be considered as a separate process from the pretraining of VLMs.
In the pretraining of CLIP, the encoders are trained to match a batch of noisy web images with their corresponding texts.
Therefore, we propose to finetune the model to match the adversarial images with corresponding texts over web-scale image-text data.
Our aim is to rethink the current standard AFT paradigm and present a new paradigm that is simpler, more intuitive and yet more effective than the standard AFT paradigm.
This section investigates the impact of the training data in more depth.


\subsection{Impact on Non-Adversarial Finetuning}

\begin{table*}[t]
    \centering
    \resizebox{\textwidth}{!}{
    \begin{tabular}{c|c|cccccccccccccc|c}
    \toprule
    \raisebox{3\height}{(\%)} &\spin{ImageNet} &\spin{CIFAR10} &\spin{CIFAR100} &\spin{STL10} &\spin{Caltech101} &\spin{Caltech256} &\spin{OxfordPets} &\spin{Flowers102} &\spin{Food101} &\spin{StanfordCars} &\spin{SUN397} &\spin{Country211} &\spin{FGVCAircraft} &\spin{EuroSAT} &\spin{DTD} &\spin{avg.} \\
    \midrule
    \color{lightgray}CLIP &\color{lightgray}59.75 & \color{lightgray}85.05 & \color{lightgray}57.18 &\color{lightgray}96.41 &\color{lightgray}86.19 &\color{lightgray}82.04
    &\color{lightgray}87.27 &\color{lightgray}65.62 &\color{lightgray}83.83 &\color{lightgray}52.13 &\color{lightgray}58.87 &\color{lightgray}15.26 &\color{lightgray}20.16 &\color{lightgray}38.32 &\color{lightgray}40.11 &\color{lightgray}62.03 \\
    
    web data &59.69 &84.31 &58.77 &95.92 &86.57 &82.42 &87.33 &65.56 &83.06 &52.94 &60.67 &15.59 &21.15 &34.58 &41.12 &62.14 \\
    IN &67.19 &84.33 &58.14 &96.52 &85.02 &82.06 &87.16 &62.29 &77.99 &46.61 &59.27 &13.69 &17.43 &35.83 &40.11 &60.46 \\
    \bottomrule
    \end{tabular}
    }
    \caption{The clean accuracy of the models finetuned on 100k image-text web data and a 100k-subset of ImageNet, respectively. Both baselines are finetuned with clean non-adversarial images.}
    \label{tab:clean_images_ft}
\end{table*}

Finetuning the model weights of pretrained VLMs, even with clean non-adversarial data, can already compromise the generalisation of the model.
We conduct a preliminary experiment to reveal the importance of following CLIP's pretraining data distribution.
Specifically, we collect 100k noisy image-text pairs from the web, and utilise them to finetune $f_\theta$ without creating adversarial images.
As a reference, we randomly sample 100 images per class from ImageNet, resulting in 100k labelled images in total.
We employ this subset of images to finetune $f_\theta$. 
For both toy datasets, we finetune for 10 epochs with the learning rate of $5e-5$.
We report the clean accuracy in \cref{tab:clean_images_ft}.
From this toy preliminary experiment, it can be seen that despite its extensive and less noisy nature, finetuning $f_\theta$ on the clean images of ImageNet already causes a slight degradation of generalisation.
In contrast, when finetuning $f_\theta$ on web-scale image-text pairs, the zero-shot performance of the model even slightly improves.
This highlights the importance of following the pretraining data distribution when modifying the model weights of VLMs, 
whereas treating finetuning as a separate process from pretraining of VLMs and modifying model weights is not an ideal choice.

\subsection{Image-Text Pairs from ImageNet}

\begin{figure}[t]
    \centering

    \begin{subfigure}{0.45\linewidth}
        \centering
        \includegraphics[width=\linewidth]{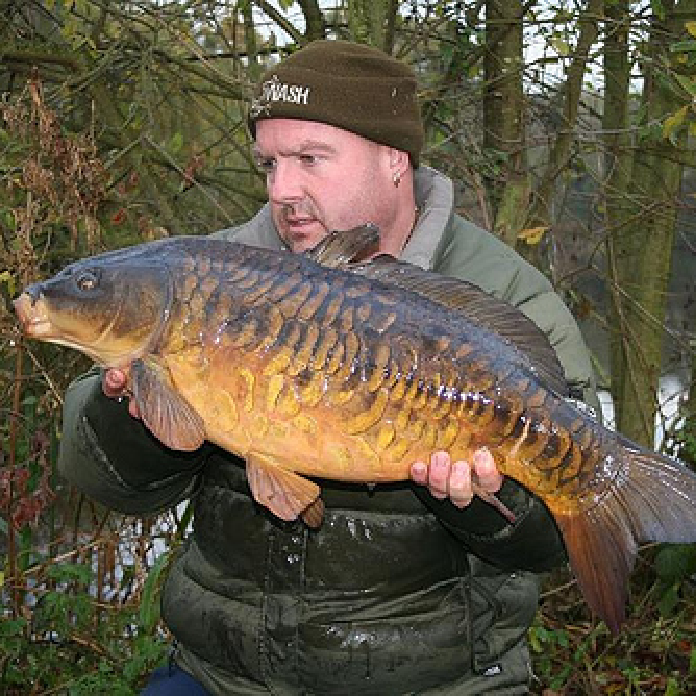}
        \caption{}
    \end{subfigure}
    \hfill
    \begin{subfigure}{0.45\linewidth}
        \centering
        \includegraphics[width=\linewidth]{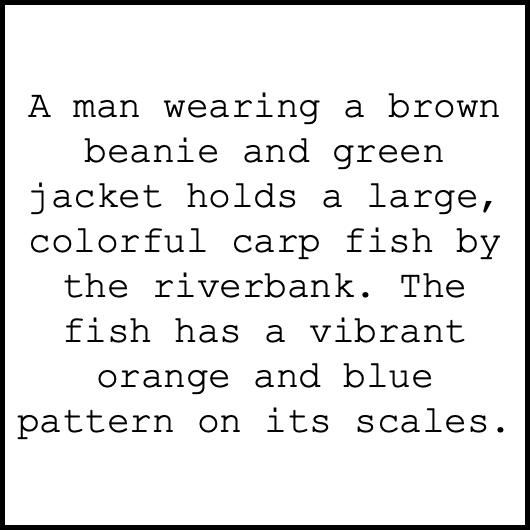}
        \caption{}
    \end{subfigure}

    \caption{An example of an image from ImageNet and its description generated by Qwen2.5-VL-3B-Instruct.}
    \label{fig:example_captioned_image}
\end{figure}

\begin{table*}[t]
    \centering
    \resizebox{\textwidth}{!}{
    \begin{tabular}{c|c|cccccccccccccc|c}
    \toprule
    \raisebox{3\height}{(\%)} &\spin{ImageNet} &\spin{CIFAR10} &\spin{CIFAR100} &\spin{STL10} &\spin{Caltech101} &\spin{Caltech256} &\spin{OxfordPets} &\spin{Flowers102} &\spin{Food101} &\spin{StanfordCars} &\spin{SUN397} &\spin{Country211} &\spin{FGVCAircraft} &\spin{EuroSAT} &\spin{DTD} &\spin{avg.} \\
    \midrule
    PMG-AFT &37.17 &39.23 &19.40 &76.79 &71.14 &59.34 &62.42 &30.69 &29.86 &14.48 &29.26 &2.59 &4.65 &12.83 &21.76 &33.89  \\
    $\textrm{AFT}_{capIN}$ &34.34 &49.71 &26.08 &82.36 &71.82 &60.37 &64.95 &34.33 &37.27 &19.28 &28.96 &2.71 &4.77 &11.12 &23.30 &36.93 \\
    $\mathrm{AdvFLYP}_{full}$ &29.47 &50.15 &25.99 &78.50 &72.47 &61.31 &59.88 &37.81 &37.53 &24.61 &34.44 &3.46 &5.73 &11.46 &24.95 &37.74  \\
    \bottomrule
    \end{tabular}
    }
    \caption{Robustness of the models finetuned on class-labelled ImageNet (PMG-AFT), VLM-captioned ImageNet ($\textrm{AFT}_{capIN}$), and our $\textrm{AdvFLYP}_{full}$, evaluated with AutoAttack ($\epsilon=1/255$).}
    \label{tab:captioned_ImageNet}
\end{table*}

In this section, we employ a generative VLM Qwen2.5-VL-3B-Instruct to generate a semantically-rich textual description for each image in ImageNet. We use the prompt of \texttt{`Describe this image with no more than 50 words'}. 
We provide a captioned example of an training image in \cref{fig:example_captioned_image}.
It can be seen that the chosen generative VLM is able to produce highly informative and coherent textual descriptions for images.
We then leverage the ImageNet dataset with the textual descriptions to perform AFT.
As in our proposed paradigm, we employ the contrastive loss with the batch size of 256, and impose regularisation on both logit and feature level. 
Results reported in \cref{tab:captioned_ImageNet} show that performing AFT on the captioned ImageNet with a contrastive loss mitigates the memorisation of the finetuning data (with a lower reported number on ImageNet), while improving the robustness across downstream datasets to a limited extent.
In comparison, $\textrm{AdvFLYP}_{full}$ outperforms the variant that performs AFT on captioned images of ImageNet, despite leveraging noisy web-scale image-text pairs, showing the 

In comparison, our $\textrm{AdvFLYP}_{full}$ paradigm consistently outperforms both baselines, showing the superiority of our simple paradigm of following the pretraining behaviour in AFT instead of treating them as separate processes.


\section{Discussion on Feature Regularisation}
\label{App_sec:feature_reg_discussion}

\citet{wang2024pre} propose logit-level regularisation on top of TeCoA, showing that it boosts the generalisation of robustness and clean accuracy across downstream datasets by penalising the logit discrepancy between the adversarial logits by the target model $f_\theta$ and adversarial logits by the frozen pretrained vision encoder $F_{\theta_0}$ (first term of \cref{eq:logit_reg}), and the logit discrepancy between adversarial logits by the target model and the clean logits by the target model (second term of \cref{eq:logit_reg}).
We conduct a trial experiment by imposing feature-level regularisation on top of TeCoA, and report the results in \cref{tab:tecoa_reg_ablation}.
The results reveal different behaviour of the TeCoA \cite{mao2023understanding} and our AdvFLYP paradigms.
Penalising discrepancy on the logit level achieves significantly large improvement over TeCoA (compare \textit{a} and \textit{b}) in terms of both downstream robustness and clean accuracy, whereas imposing regularisation on the feature level brings marginal effects (compare \textit{a} and \textit{c}).
In contrast, as reported in \cref{tab:regularisation_ablation} in the main paper, logit- and feature-level penalties play different roles, with $\mathcal{L}_{logit}$ benefiting transferability of robustness gains across downstream datasets and $\mathcal{L}_{feat}$ facilitating preservation of zero-shot capabilities on clean images. 
We believe there are two main reasons. Firstly, the prior TeCoA paradigm caters to the classification task, where the logit is the key element.
Additionally, they produce adversarial images that maximise the cross-entropy loss \wrt a pre-defined set of categories. This may not cause a significant shift of the embeddings in the latent space.
In contrast, our AdvFLYP creates adversarial images based on noisy web images \wrt their texts, which can result in considerable embedding shifts. Finetuning $f_\theta$ with these distorted adversarial embeddings can contaminate the vision encoder.
Therefore, penalising the deviation of image features with $\mathcal{L}_{feat}$ is crucial for retaining CLIP's zero-shot performance on clean images.

\begin{figure}[t]
    \centering
    \begin{subfigure}{0.3\linewidth}
        \centering
        \includegraphics[width=\linewidth]{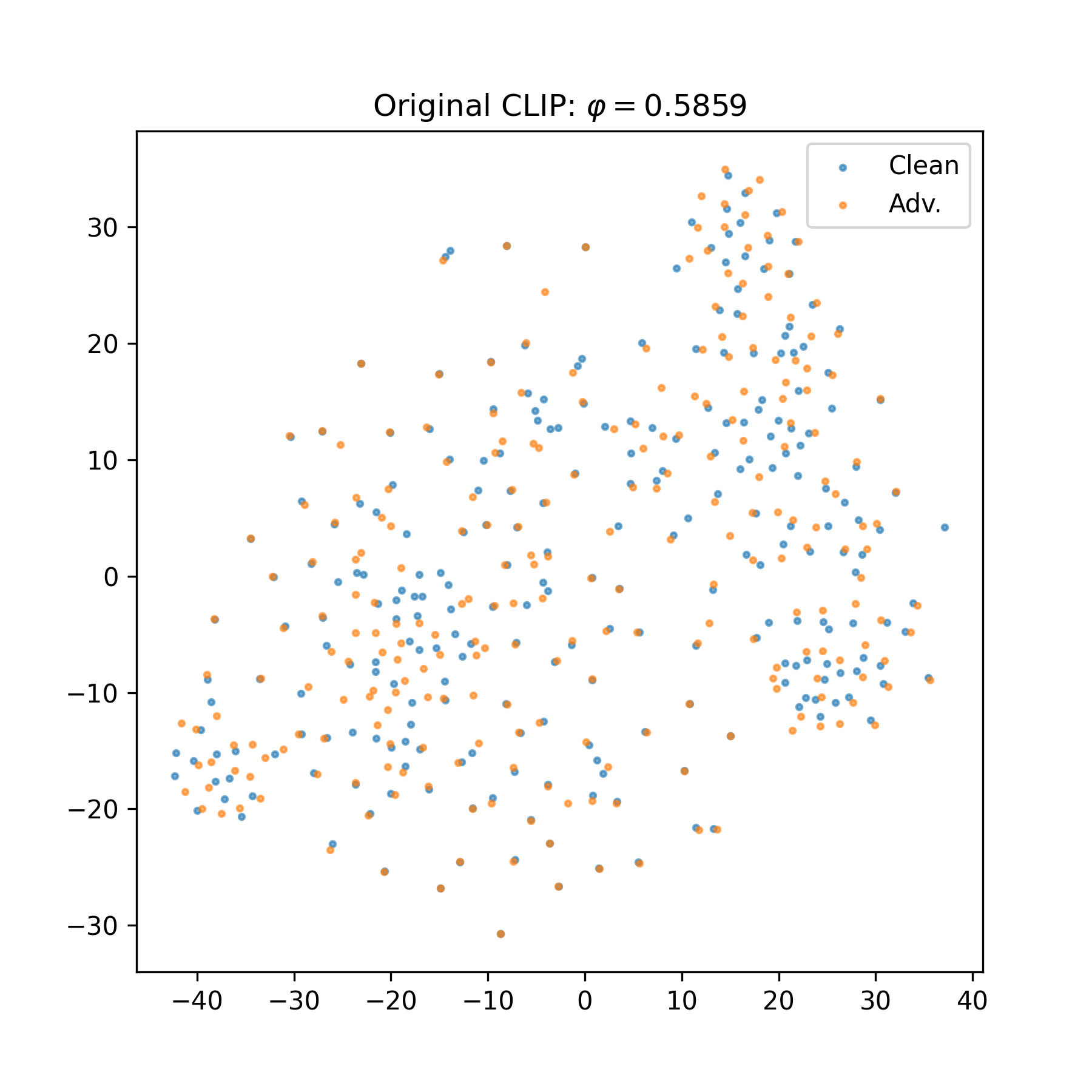}
        \caption{Original CLIP}\label{clip_orig_tsne}
    \end{subfigure}
    \hfill
    \begin{subfigure}{0.3\linewidth}
        \centering
        \includegraphics[width=\linewidth]{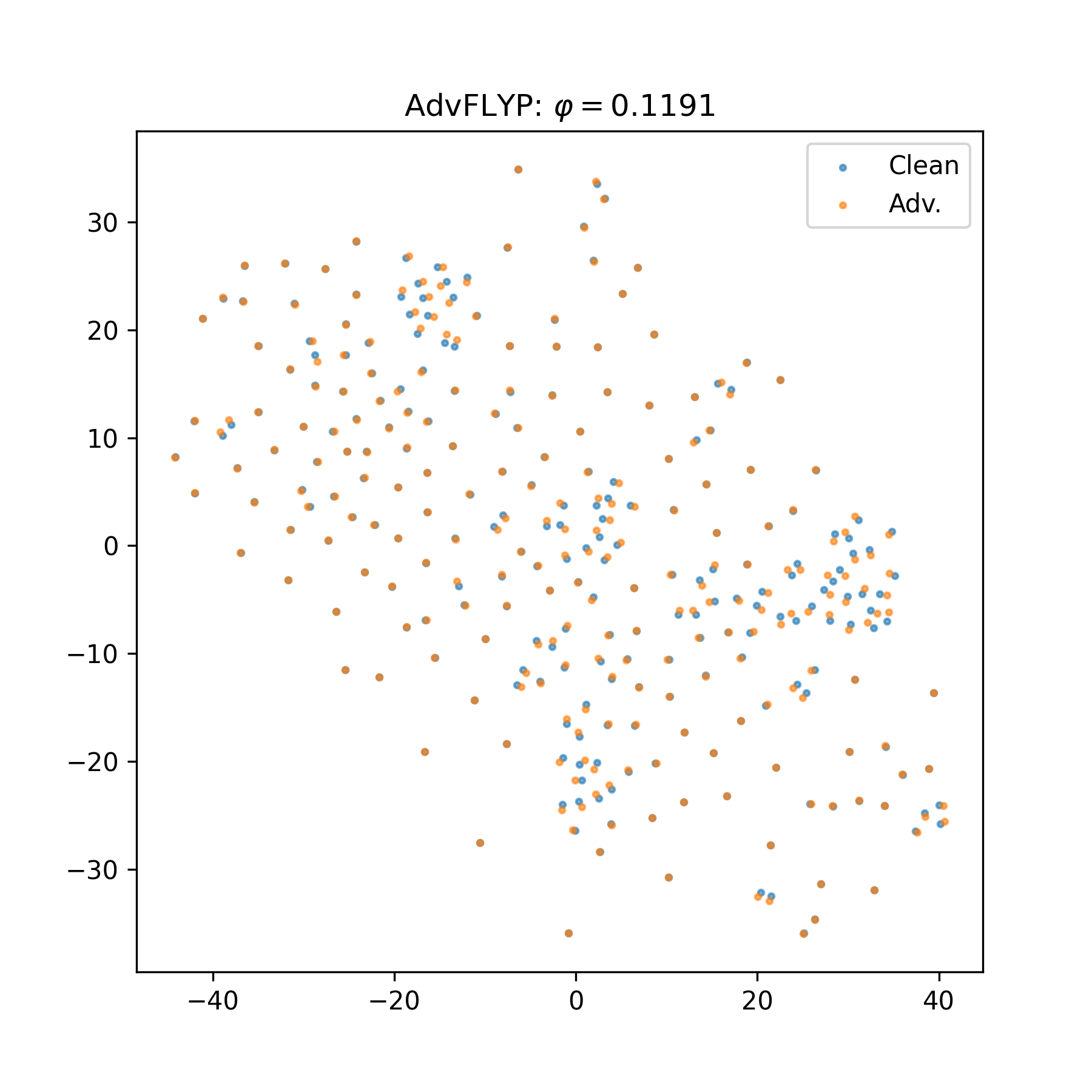}
        \caption{AdvFLYP}\label{advflyp_tsne}
    \end{subfigure}
    \hfill
    \begin{subfigure}{0.3\linewidth}
        \centering
        \includegraphics[width=\linewidth]{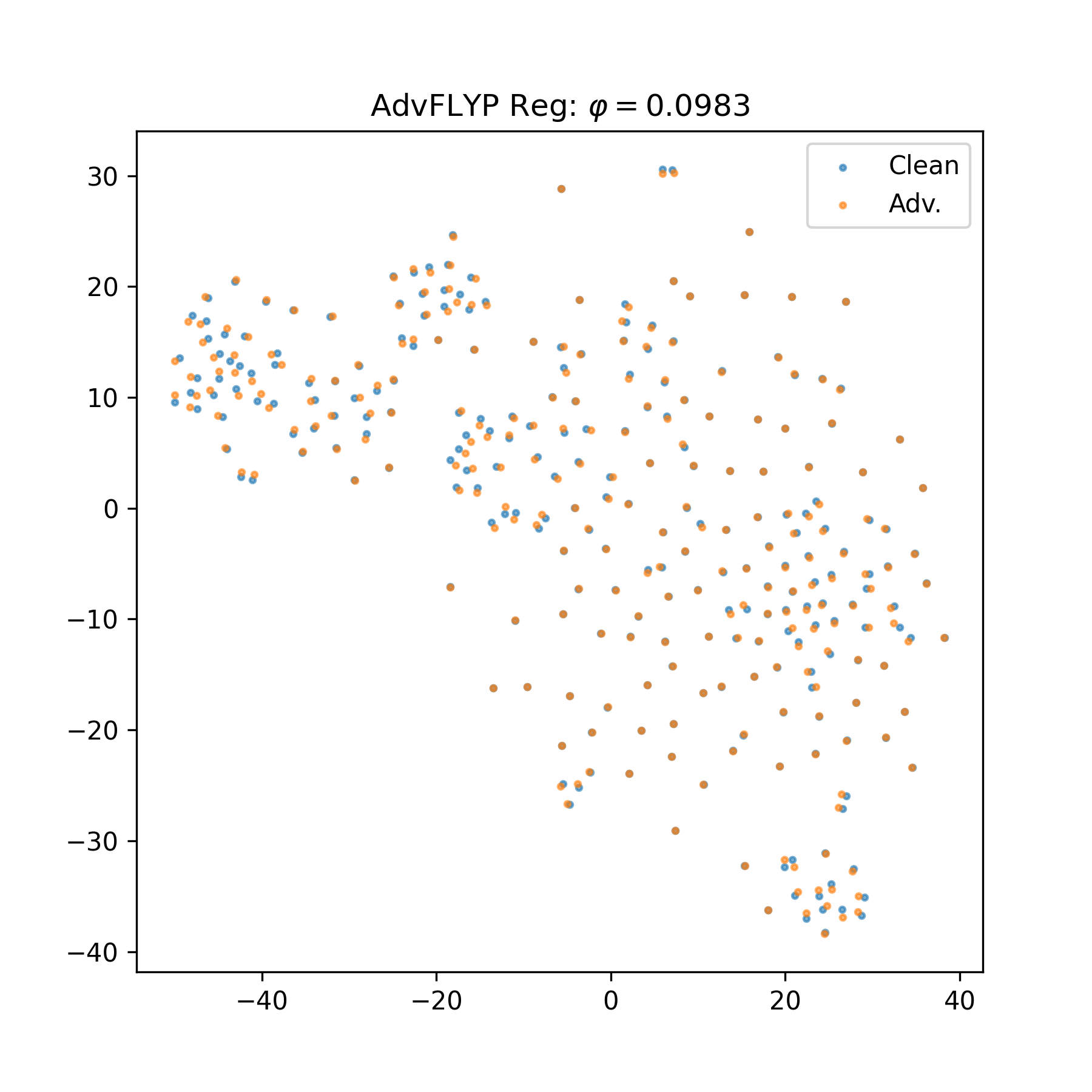}
        \caption{$\textrm{AdvFLYP}_{full}$}\label{advflyp_full_tsne}
    \end{subfigure}
    \caption{t-SNE visualisation of adversarial and clean image features in the latent space.}\label{fig:tsne}
\end{figure}

To further investigate the effects of imposing regularisation over AdvFLYP, 
we analyse the cosine deviation of clean and adversarial features of AdvFLYP and the regularised variant $\textrm{AdvFYLP}_{full}$.
We define the cosine deviation as follows:
\begin{equation}
    \varphi=\arccos
    \frac{
    f_\theta(x)^\intercal f_\theta(x+\delta)
    }
    {\| f_\theta(x) \| \cdot
    \| f_\theta(x+\delta) \|
    }
\end{equation}
where a larger $\varphi$ indicates greater cosine deviation of adversarial image features from their clean counterparts in the latent space, and vice versa.
Specifically, we sample 256 images from ImageNet and employ the t-SNE algorithm visualise the adversarial and clean image features for AdvFLYP and $\textrm{AdvFLYP}_{full}$. 
As can be seen from \cref{fig:tsne}, adversarial image features deviate significantly from their clean counterparts in the latent space of the original CLIP, with average cosine deviation at $\varphi=0.5859$, whereas AdvFLYP and its regularised variant $\textrm{AdvFLYP}_{full}$ effectively mitigating such deviation, as evidenced by the narrowed gap between adversarial and clean features. Imposing regularisation ($\textrm{AdvFLYP}_{full}$) further mitigates the cosine deviation as $\varphi$ is decreased to 0.0983, in comparison to $\varphi=0.1191$ achieved by AdvFLYP with no regularisation terms.

\end{document}